\documentclass[10pt,twocolumn,letterpaper]{article}

\usepackage{cvpr}
\usepackage{times}
\usepackage{epsfig}
\usepackage{graphicx}
\usepackage{amsmath}
\usepackage{amssymb}

\usepackage[breaklinks=true,bookmarks=false]{hyperref}

\usepackage{subfigure,multirow}
\usepackage{algorithm}
\usepackage{algorithmic}
\newtheorem{proposition}{Proposition}

\newtheorem{definition}{Definition}

\cvprfinalcopy 

\def\cvprPaperID{9235} 

\ifcvprfinal\fi
\begin{document}

\title{ Alleviation of Gradient Exploding in GANs: Fake Can Be Real}

\author{
Song Tao, Jia Wang\thanks{Corresponding author}\\
Department of Electronic Engineering\\
Shanghai Jiao Tong University\\
{\tt\small \{taosong, jiawang\}@sjtu.edu.cn}
}


\maketitle
\thispagestyle{empty}

\begin{abstract}
   In order to alleviate the notorious mode collapse phenomenon in generative adversarial networks (GANs), we propose a novel training method of GANs in which certain fake samples are considered as real ones during the training process. This strategy can reduce the gradient value that generator receives in the region where gradient exploding happens. We show the process of an unbalanced generation and a vicious circle issue resulted from gradient exploding in practical training, which explains the instability of GANs. We also theoretically prove that gradient exploding can be alleviated by penalizing the difference between discriminator outputs and fake-as-real consideration for very close real and fake samples. Accordingly, Fake-As-Real GAN (FARGAN) is proposed with a more stable training process and a more faithful generated distribution. Experiments on different datasets verify our theoretical analysis.
\end{abstract}

\section{Introduction}

In the past few years, Generative Adversarial Networks (GANs) \cite{goodfellow2014generative} have been one of the most popular topics in generative models and achieved great success in generating diverse and high-quality images \cite{brock2018large,karras2019style,donahue2019large}. GANs can be expressed as a zero-sum game between discriminator and generator. When a final theoretical equilibrium is achieved, discriminator can never distinguish between real and fake generated samples. However, we show that a theoretical equilibrium actually can be seldom realized in practice with only discrete finite samples in datasets during the training process.

Although GANs have achieved remarkable progress, numerous researchers have tried to improve the performance of GANs from various aspects \cite{pmlr-v70-arjovsky17a,nowozin2016f,gulrajani2017improved,miyato2018spectral}, because of the inherent problem in GAN training, such as instability and mode collapse. \cite{Arora0LMZ17} showed that a theoretical generalization guarantee does not be provided with the original GAN objective and analyzed the generalization capacity of neural network distance. The author argued that for a low capacity discriminator, it can not provide generator enough information to fit the target distribution owing to lack of ability to detect mode collapse. \cite{thanh2019improving} argued that poor generation capacity in GANs comes from the discriminators trained on finite training samples resulting in overfitting to real data samples and gradient exploding when generated datapoints approach real ones. As a result, \cite{thanh2019improving} proposed a zero-centered gradient penalty on linear interpolations between real and fake samples to improve generalization capability and prevent mode collapse resulted from gradient exploding. Recent work \cite{wu2019generalization} further studied generalization from a new perspective of privacy protection.

In this paper, we focus on mode collapse resulted from gradient exploding studied in \cite{thanh2019improving} and achieve a better generalization with a much more stable training process. Our contributions are as follows:
\begin{enumerate}
\vspace{-0.1cm}
\item {We explain the generation process of an unbalanced distribution in GAN training, which becomes more and more serious as training progresses owing to the existence of the vicious circle issue resulted from gradient exploding.}
\vspace{-0.1cm}
\item {We prove that the gradient exploding issue can be effectively alleviated by difference penalization for discriminator between very close real and fake samples and fake-as-real consideration where gradient exploding happens.}
\vspace{-0.1cm}
\item {We propose a novel GAN training method by considering certain fake samples as real ones (FARGAN) according to discriminator outputs in a training minibatch to effectively prevent the unbalanced generation. Experiments on synthetic and real world datasets verify that our method can stabilize training process and achieve a more faithful generated distribution.}
\end{enumerate}

In the sequel, we use the terminologies of generated samples (datapoints) and fake samples (datapoints) indiscriminately. Tab. \ref{tab:notations} lists some key notations used in the rest of the paper.

\newcommand{\tabincell}[2]{
}
\begin{table*}[!t]
  \centering
  \caption{NOTATIONS}
  \label{tab:notations}
  \begin{tabular}{ll}
    \\[-2mm]
    \hline
    \\[-2mm]
    {\bf \small Symbol}&\qquad {\bf\small Meaning}\\
    \hline
    \vspace{1mm}\\[-3mm]
    $p_r$      &   the target dsitribution\\
    \vspace{1mm}
    $p_g$          &  the model distribution\\
     \vspace{1mm}
     $D$          &  the discriminator with sigmoid function in the last layer\\
    \vspace{1mm}
     $D_0$         & the discriminator with sigmoid function in the last layer removed\\
     \vspace{1mm}
    $D_r=\{{x}_1,\cdots,{x}_n\}$  &   the set of $n$ real samples\\
     \vspace{1mm}
    $D_g=\{{y}_1,\cdots,{y}_m\}$  &   the set of $m$ generated samples\\
    \vspace{1mm}
    $D_{\text{FAR}}=\{\widetilde{y}_1,\cdots,\widetilde{y}_{N_0}\}$  &  the set of $N_0$ generated samples considered as real\\
    \hline
  \end{tabular}
\end{table*}

\section{Related work}
\textbf{Instability.} GANs have been considered difficult to train and often play an unstable role in training process \cite{SalimansGZCRCC16}. Various methods have been proposed to improve the stability of training. A lot of works  stabilized training with well-designed structures \cite{radford2015unsupervised,karras2018progressive,ZhangGMO19,chen2018on} and utilizing better objectives \cite{nowozin2016f,zhao2016energy,pmlr-v70-arjovsky17a,mao2017least}. Gradient penalty to enforce Lipschitz continuity is also a popular direction to improve the stability including \cite{gulrajani2017improved,petzka2018on,roth2017stabilizing,qi2017loss}.
From the theoretical aspect, \cite{nagarajan2017gradient} showed that GAN optimization based on gradient descent is locally stable and \cite{mescheder2018training} proved local convergence for
simplified zero-centered gradient penalties under suitable
assumptions. For a better convergence, a two time-scale update rule (TTUR) \cite{heusel2017gans} and exponential moving averaging (EMA) \cite{yazc2018the} have also been studied.

\textbf{Mode collapse.} Mode collapse is another persistent essential problem for the training of GANs, which means lack of diversity in the generated samples. The generator may sometimes fool the discriminator by producing
a very small set of high-probability samples from the data
distribution. Recent work \cite{Arora0LMZ17,arora2018do} studied the generalization capacity of GANs and showed that the model distributions learned by GANs do miss a significant number of modes. A large number of ideas have been proposed to prevent mode collapse. Multiple generators are applied in \cite{Arora0LMZ17,ghosh2018multi,hoang2018mgan}
to  achieve a more faithful distribution. Mixed samples are considered as the inputs of discriminator in \cite{LinKFO18,LucasTOV18} to convey information on diversity. Recent work \cite{he2018bayesian} studied mode collapse from  probabilistic treatment and \cite{yamaguchi2018distributional,dieng2019prescribed} from the entropy of distribution.

\section{Background}
\vspace{-0.1cm}
In the original GAN \cite{goodfellow2014generative}, the discriminator D maximizes the following objective:
\begin{equation}
\mathcal{L}= {E}_{{x}\sim p_r}[\log(D({x}))]+{E}_{{y}\sim p_g}[\log(1-D({y}))],
\label{eq:1}
\end{equation}
and to prevent gradient collapse, the generator G in Non-Saturating GAN (NSGAN) \cite{goodfellow2014generative} maximizes
\begin{equation}
\mathcal{L}_G= {E}_{{y}\sim p_g}[\log(D({y}))],
\label{eq:2}
\end{equation}
where $D$ is usually represented by a neural network.
\cite{goodfellow2014generative} showed that the optimal discriminator $D$ in Eqn.\ref{eq:1} is $D^*({v})=\frac{p_r({v})}{p_r({v})+p_g({v})}$
for any ${v}\in supp(p_r) \cup supp(p_g)$. As training progresses, $p_g$ will be pushed closer to $p_r$. If G and D are given enough capacity, a global equilibrium is reached when $p_r=p_g$, in which case the best strategy for D on $supp(p_r) \cup supp(p_g)$ is just to output $\frac{1}{2}$ and the optimal value for Eqn.\ref{eq:1} is $2\log(\frac{1}{2})$.

With finite training examples in training dataset $D_r$ in practice, we empirically use $\frac{1}{n}\sum_{i=1}^{n}\log(D({x_i}))$ to estimate ${E}_{{x}\sim p_r}[\log(D({x}))]$ and $\frac{1}{m}\sum_{i=1}^{m}[1-\log(D({y_i}))]$ to estimate ${E}_{{y}\sim p_g}[\log(1-D({y}))]$, where ${x_i}$, ${y_i}$ is from $D_r$ and generated dataset $D_g$, respectively.

Mode collapse in generator is attributed to gradient exploding in discriminator, according to \cite{thanh2019improving}. When a fake datapoint ${y}_0$ is pushed to a real datapoint ${x}_0$ and if  $|D({x_0})-D({y_0})|\geq\epsilon$ is satisfied, the absolute value of directional derivative of D in the direction ${\mu}={x_0}-{y_0}$ will approach infinity:
\begin{eqnarray}
|(\nabla_{{\mu}}D)_{{x_0}}|&=&\lim_{{y_0}\stackrel{{\mu}}{\rightarrow}{x_0}}\frac{|D({x_0})-D({y_0})|}{||{x_0}-{y_0}||}
\nonumber \\
&
\geq &\lim_{{y_0}\stackrel{{\mu}}{\rightarrow}{x_0}}\frac{\epsilon}{||{x_0}-{y_0}||}=\infty,
\end{eqnarray}
in which case the gradient norm of discriminator at ${y_0}$, $||\nabla_{{y_0}} D({y_0})||$, is equivalent to $|(\nabla_{{\mu}}D)_{{x_0}}|$ and gradient explodes.
Since $\nabla_{{y_0}} D({y_0})$ outweighs gradients towards other modes in a training minibatch, gradient exploding at datapoint ${y_0}$ will move multiple fake datapoints towards ${x_0}$ resulting in mode collapse.

\begin{figure*}
  \centering
  \subfigure[]{\includegraphics[width=1.3in]{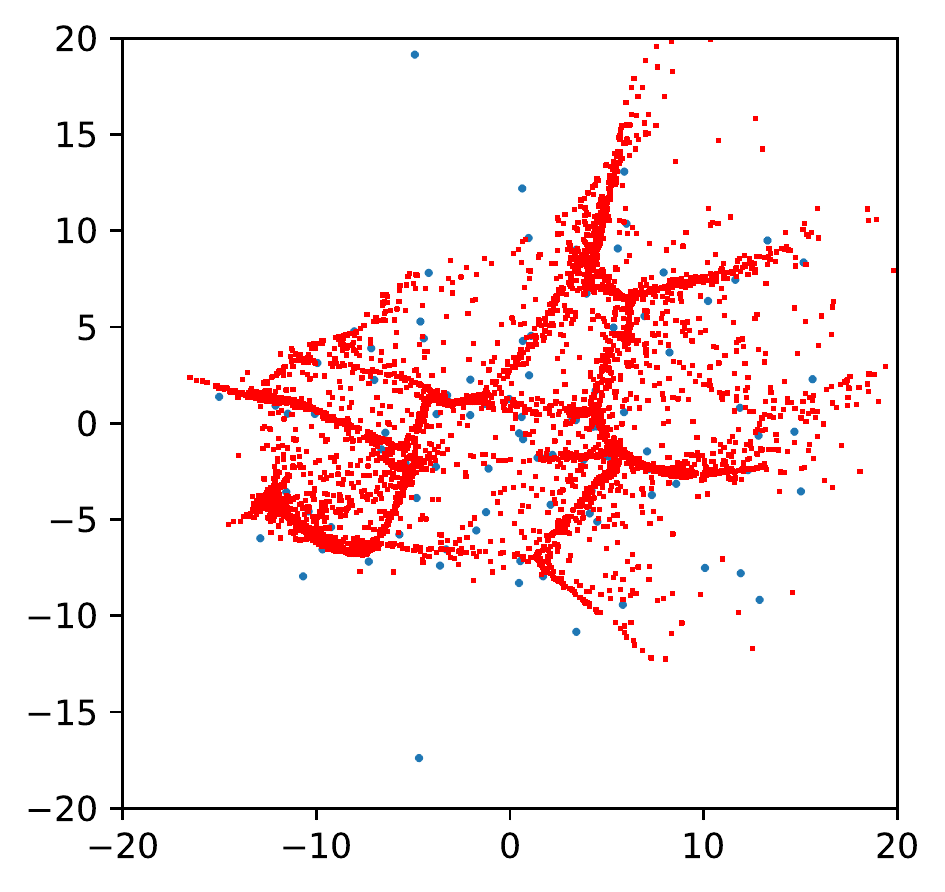}}
  \hspace{.1in}
    \centering
  \subfigure[]{\includegraphics[width=1.3in]{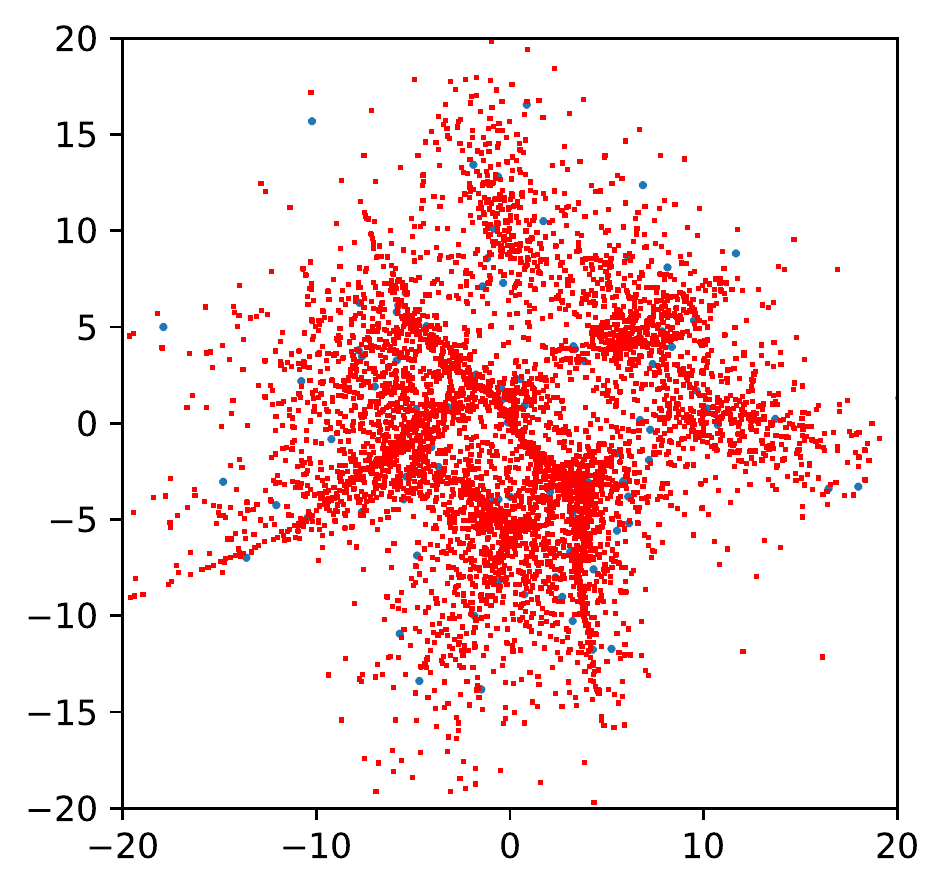}}
  \hspace{.1in}
   \centering
  \subfigure[]{\includegraphics[width=1.3in]{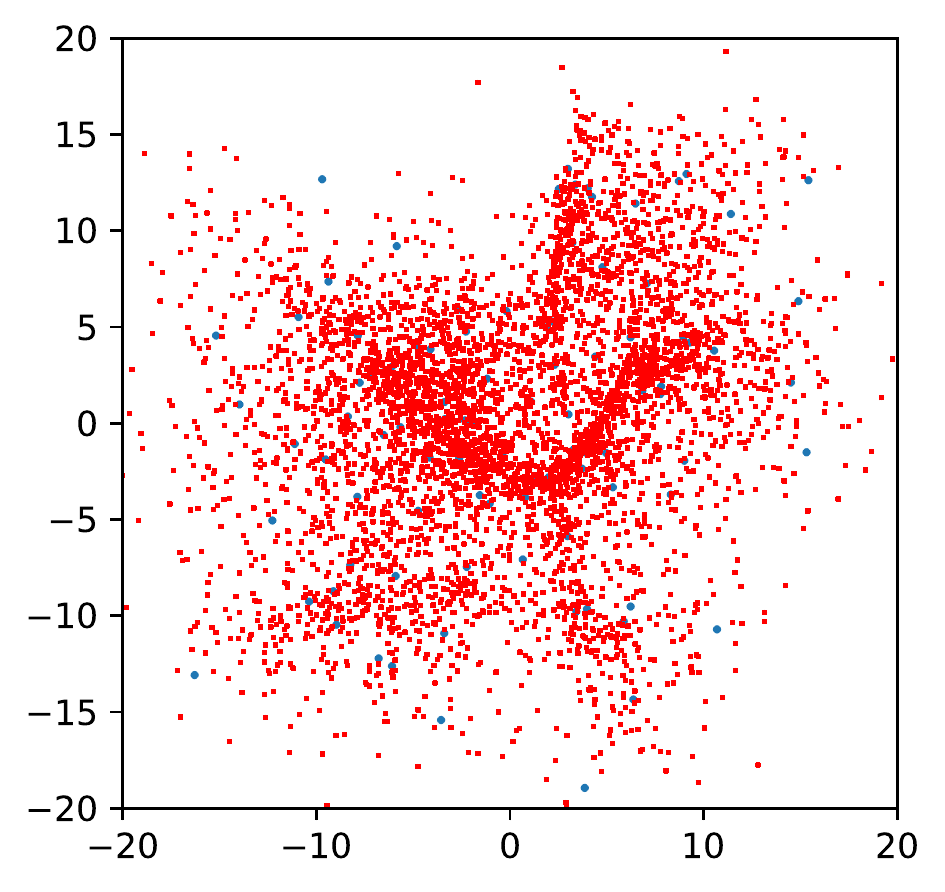}}
  \hspace{.1in}
   \centering
  \subfigure[]{\includegraphics[width=1.3in]{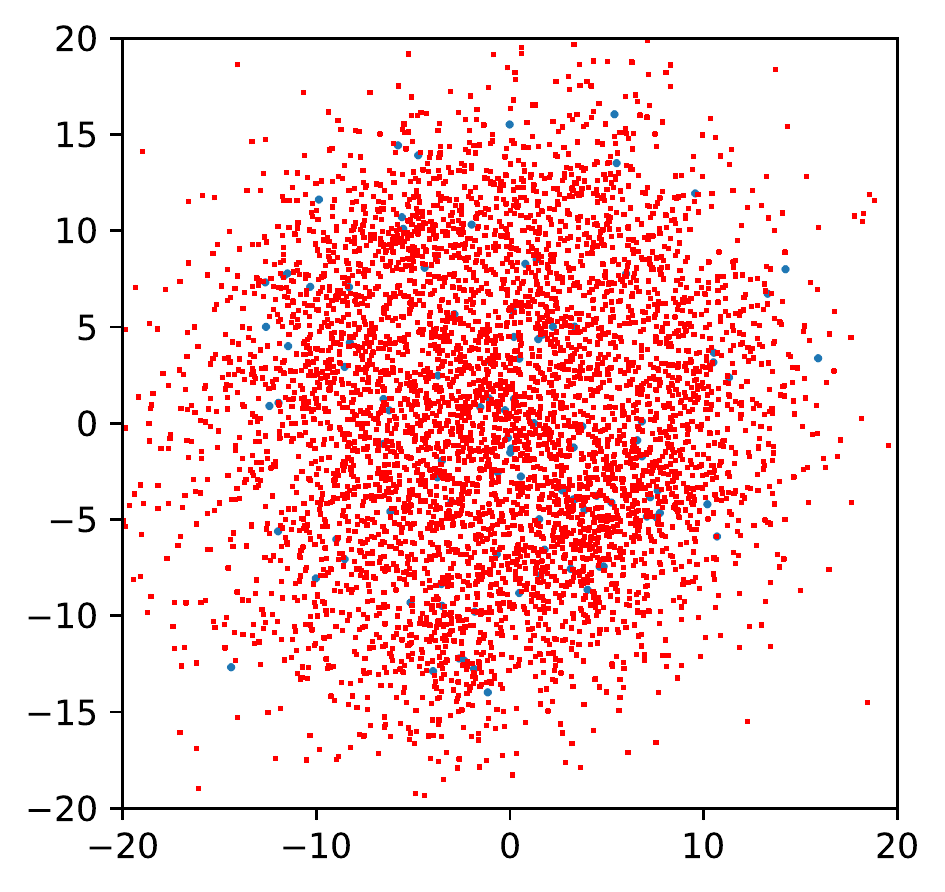}}
  \centering
  \subfigure[]{\includegraphics[width=1.3in]{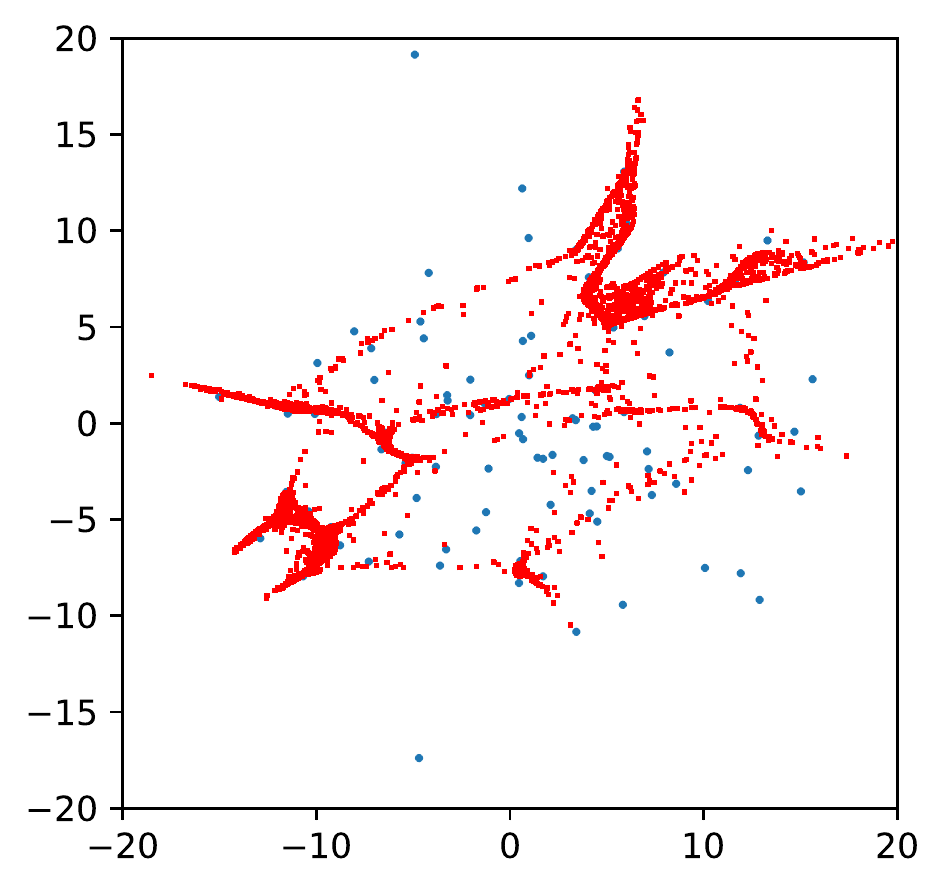}}
  \hspace{.1in}
    \centering
  \subfigure[]{\includegraphics[width=1.3in]{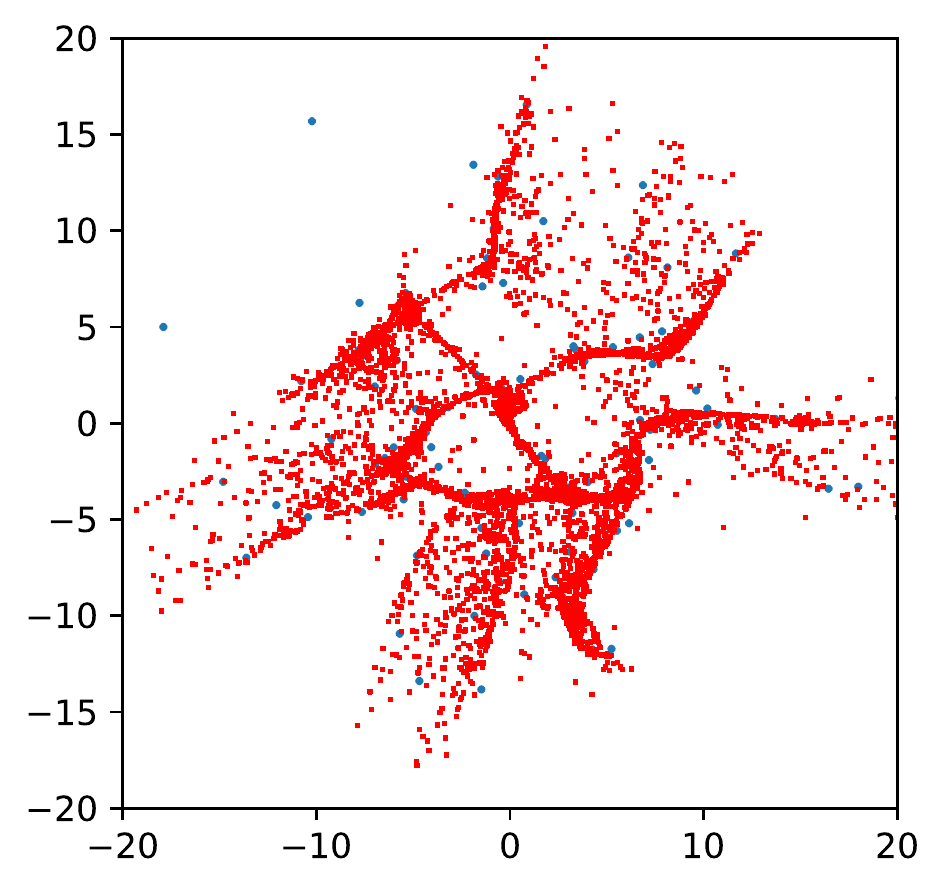}}
  \hspace{.1in}
   \centering
  \subfigure[]{\includegraphics[width=1.3in]{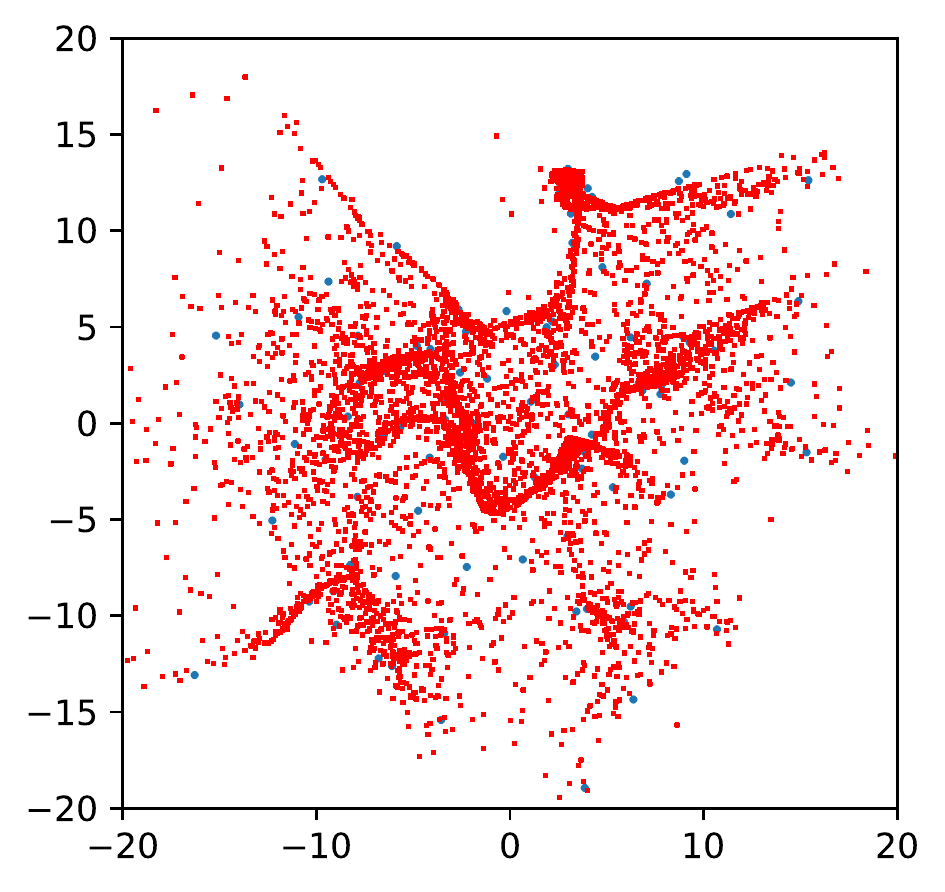}}
  \hspace{.1in}
   \centering
  \subfigure[]{\includegraphics[width=1.3in]{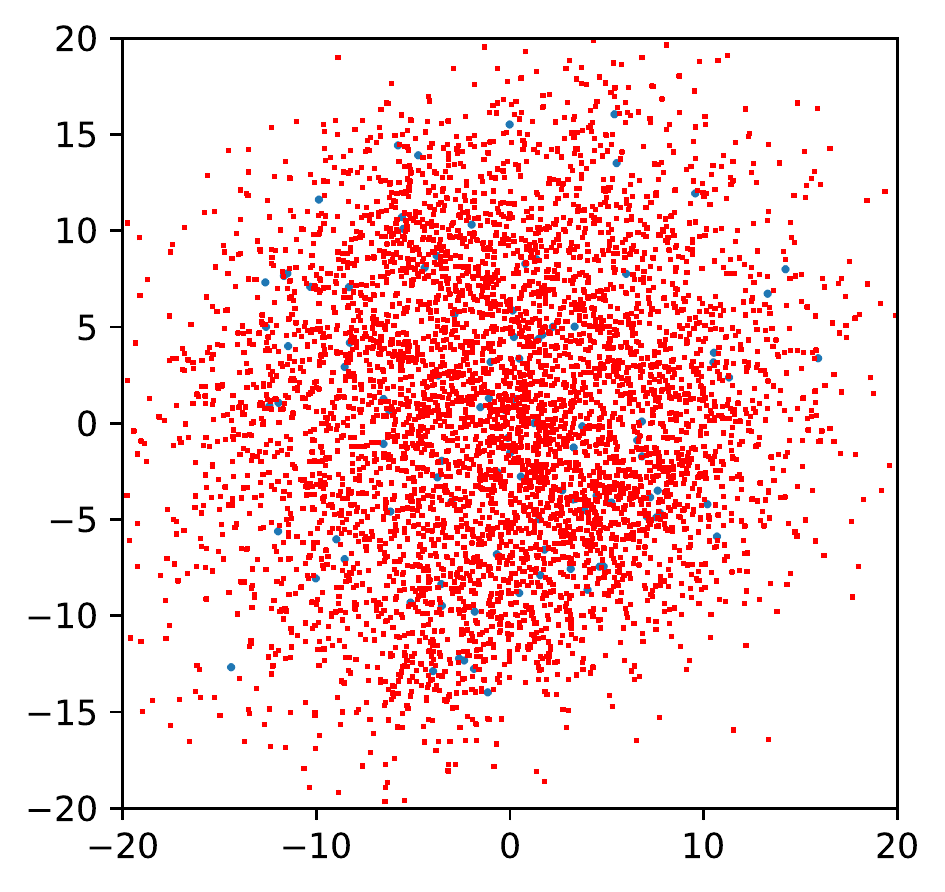}}
  \caption{Results on finite samples from a Gaussian distribution of GANs trained with different gradient penalties and our method. Blue datapoints represent real samples and red datapoints represent generated samples. (a)(e) NSGAN with no GP, iter. 100k and 200k. (b)(f) NSGAN-0GP-sample, iter. 100k and 200k. (c)(g) NSGAN-0GP-interpolation, iter. 100k and 200k. (d)(h) NSGAN-0GP-sample with our method, iter. 100k and 200k. }
  \label{fig:toy1}
\end{figure*}

\section{Unbalanced Generation}
\label{sec:1}
Theoretically, discriminator outputs a constant $\frac{1}{2}$ when a global equilibrium is reached. However in practice, discriminator can often easily distinguish between real and fake samples \cite{goodfellow2014generative,pmlr-v70-arjovsky17a}. Because the target distribution $p_r$ is unknown for discriminator, discriminator will always consider training samples in $D_r$ as real while generated samples in $D_g$ as fake. Even when the generated distribution $p_g$ is equivalent to the target distribution $p_r$, $D_r$ and $D_g$ is disjoint with probability $1$ when they are sampled from two continuous distributions respectively (Proposition 1 in \cite{thanh2019improving}). In this case, actually $D_g$ is pushed towards samples in $D_r$. We will explain specifically the generation process of an unbalanced distribution that deviates from $p_r$.

\begin{definition}
For $x_0\in D_r,y_0\in D_g$, $\{x_0,y_0\}$ is a $\delta~\text{close}$ pair if ${y_0}\in \mathcal{N}^\delta({x_0})=\{{y_0}: d({x_0},{y_0})\leq\delta, 0<\delta\ll d({x_i},{x_j}), \forall x_i,x_j \in D_r\}$. Additionally,
$x_0$ is called an overfitting source in a close pair $\{x_0,y_0\}$.
\end{definition}

During the process of $D_g$ approaching $D_r$, multiple overfitting sources will appear.
The following proposition shows that the optimal empirical discriminator does not give equal outputs between the corresponding real and fake samples for all close pairs.

\begin{proposition}
If overfitting sources exist, an empirical discriminator satisfying $D(x_0)-D(y_0)\geq\epsilon$ on a close pair $\{x_0,y_0\}$ can be easily constructed as a MLP with only $\mathcal{O}(2\dim(x))$ parameters.
\label{pro:1}
\end{proposition}

See Appendix A for the detailed proof.
The discriminators used in practice usually contains hundreds of millions parameters, which are much more powerful than the discriminator we constructed above. Although \cite{thanh2019improving} constructed a discriminator to distinguish all samples between $D_r$ and $D_g$, they use much more parameters which are comparable to that used in practice and we needn't distinguish all samples but only a close pair $\{x_0,y_0\}$.

From Eqn.\ref{eq:2}, the gradient norm generator receives from discriminator at $y_0$ for a close pair $\{x_0,y_0\}$ can be computed as
\begin{eqnarray}
||\nabla_{{y_0}}\mathcal{L}_G({{y_0}})||\!=\!
\frac{1}{D({y_0})}\!\lim_{{y_0}\stackrel{{\mu}}{\rightarrow}{x_0}}\!\!\frac{|D({x_0})-D({y_0})|}{||{x_0}-{y_0}||}.
\end{eqnarray}
When $D({x_0})-D({y_0})\geq\epsilon$ is satisfied and $\{x_0,y_0\}$ happens to be a close pair, the gradient of generator at $y_0$ explodes and outweighs the gradients towards other modes excessively. Fake samples will be moved in the direction ${\mu}={x_0}-{y_0}$ and especially other fake samples in a minibatch will not be moved towards the corresponding modes, making an unbalanced generation visible.
See the generated results on a Gaussian dataset of the original GAN in Fig.~\ref{fig:toy1}a, \ref{fig:toy1}e. The generated distribution neither covers the target Gaussian distribution nor fits all the real samples in $D_r$.

\section{Gradient Alleviation}
In this section, we search for ways of alleviating the gradient exploding issue to achieve a more faithful generated distribution. For the simplicity of analysis, we extract sigmoid function $\sigma$ from the last layer of $D$, i.e. $D(\cdot)=\sigma(D_0(\cdot))$. The gradient norm of generator at $y_0$ for a close pair $\{x_0,y_0\}$ can be rewritten as
\begin{eqnarray}
||\nabla_{{y_0}}\mathcal{L}_G({{y_0}})||\!=\!
\sigma(-D_0({y_0}))\!
\lim_{{y_0}\stackrel{{\mu}}{\rightarrow}{x_0}}\!\!\frac{|D_0({x_0})-D_0({y_0})|}{||{x_0}-{y_0}||}.
\label{eq:6}
\end{eqnarray}
Consider the scenario in which ${x_0}$, in a set of $n$ real samples, is an overfitting source for $\{{y_1},{y_2},\cdots,{y_{m_0}}\}$, in a set of $m$ generated samples, i.e., $\{x_0,y_i\}, i=1,\cdots,m_0$ are close pairs.
We are specially interested in the outputs of the optimal discriminator at ${x_0}$ and $\{{y_1},{y_2},\cdots,{y_{m_0}}\}$. For simplicity, we make the assumption that the outputs of discriminator at these interested points are not affected by other samples in $D_r$ and $D_g$. We also assume discriminator has enough capacity to achieve the optimum in this local region.

\subsection{Difference Penalization }
\label{sec:2}
We first consider penalizing the $L_2$ norm of the output differences on close pairs, resulting in the following empirical discriminator objective:
\begin{align}\nonumber
\mathcal{L}_{\text{DP}}=&\frac{1}{n}\left[\log \sigma(D_0(x_0))+\sum_{i=1}^{n-1}\log\sigma(D_0(x_i))\right]\\\nonumber
&+\frac{1}{m}\left[\sum_{i=1}^{m_0}\log(1-\sigma(D_0({y_i})))\right. \\\nonumber
&\quad+\left.\sum_{i=m_0+1}^{m}\log(1-\sigma(D_0({y_i})))\right]\\\nonumber
&-\frac{k}{m_0}\sum_{i=1}^{m_0}(D_0({x_0})-D_0({y_i}))^2\\\label{eq:3}
=&C_1+\frac{1}{n}f(D_0({x_0}),D_0({y_1}),\cdots,D_0({y_{m_0}})),
\end{align}
where $k$ is the weight of the $L_2$ norms and $C_1$ is an inconsequential term. Denoting $D_0({x_0})$ as $\xi_0$ and $D_0({y_i})$ as $\xi_i$, $i=1,\cdots,m_0$, the interested term $f(\xi_0,\xi_1,\cdots,\xi_{m_0})$ in Eqn.\ref{eq:3} is
\begin{equation}
f\!=\!\log\sigma(\xi_0)\!+\!\frac{n}{m}\sum_{i=1}^{m_0}\log(1\!-\!\sigma(\xi_i))\!-\!\frac{nk}{m_0}\sum_{i=1}^{m_0}(\xi_0\!-\!\xi_i)^2.
\end{equation}

\begin{proposition}
Assume that $\{\xi^*_0,\cdots,\xi^*_{m_0}\}$ achieves the maximum of $f(\xi_0,\xi_1,\cdots,\xi_{m_0})$. Then
with $k$ increasing, $\sigma(-\xi_i^*)(\xi_0^*-\xi_i^*)$ decreases, and, with $m_0$ increasing, $\sigma(-\xi_i^*)(\xi_0^*-\xi_i^*)$ increases, $\forall i=1,\cdots,m_0$.
\label{pro:2}
\end{proposition}

See Appendix B for the detailed proof.
Hence, the gradient norm of generator in this local region decreases with the weight $k$ of difference penalization increasing, while increases with the number of close pairs $m_0$ increasing from Eqn.\ref{eq:6}.

{\bf Gradient penalty.} Actually in practice, it is hard to find close pairs to make the corresponding difference penalization. If we directly penalize the $L_2$ norm of $D_0({x_i})-D_0({y_i})$, the gradient norm at ${y_i}$ may get even larger when $\{{x_i},{y_i}\}$ is not a close pair. Considering $D_0({y_i})>D_0({x_i})$, which could happen when the number of close pairs at ${x_i}$ is larger than that at ${y_i}$, direct penalization will make $D_0({y_i})$ lower and further the gradient norm at ${y_i}$ larger from Eqn.\ref{eq:6}.
Thus in practice we could enforce a zero-centered gradient penalty of the form $||(\nabla D_0)_{{v}}||^2$ to stabilize the discriminator output for close pairs, where ${v}$ can be real or fake samples. Although far from perfection, Fig.~\ref{fig:toy1}b, \ref{fig:toy1}f generate more faithful results compared with Fig.~\ref{fig:toy1}a, \ref{fig:toy1}e with no gradient penalty added.

To prevent gradient exploding, \cite{thanh2019improving} proposed another zero-centered gradient penalty of the form $||(\nabla D_0)_{{v}}||^2$, where ${v}$ is a linear interpolation between real and fake samples. However, we consider it's not a very efficient method to fill the gap here. To begin with, the result of interpolation may not lie in $supp(p_r)\cup supp(p_g)$. Furthermore, for arbitrary pair of real and fake samples, the probability that linear interpolation between them lies where close pairs exist is close to $0$ especially for high-dimensional situations.

{\bf Vicious circle.} Gradient exploding near overfitting source $x_0$ results in multiple fake samples moved towards $x_0$. Then more close pairs results in a more serious gradient exploding issue, forming a vicious circle. It partly explains the instability of GAN training process that especially during the later stage of training, similar generated samples are seen. Compared with Fig.~\ref{fig:toy1}a, \ref{fig:toy1}b, \ref{fig:toy1}c at iter.100k, Fig.~\ref{fig:toy1}e, \ref{fig:toy1}f, \ref{fig:toy1}g at iter.200k have a more unbalanced generation and more similar samples are generated as training progresses.

\subsection{Fake-as-Real Consideration}

Based on discussions above, we add a fake-as-real consideration on $m_0$ fake samples $\{{y_1},{y_2},\cdots,{y_{m_0}}\}$, resulting in the following empirical discriminator objective:
\begin{eqnarray}
\mathcal{L}_{\text{FAR}}\!\!\!&=&\!\!\! \mathcal{L}_{\text{DP}}+\lambda\sum_{i=1}^{m_0}\log\sigma(D_0({y_i}))
\nonumber \\
\!\!\!&=&\!\!\!C_2+\frac{1}{n}h(\xi_0,\xi_1,\cdots,\xi_{m_0}),
\label{eq:4}
\end{eqnarray}
where $\lambda$ is the weight of considering fake as real and $C_2$ is an inconsequential term. The interested term $h(\xi_0,\xi_1,\cdots,\xi_{m_0})$ in Eqn.\ref{eq:4} is
\begin{equation}
\label{eq:9}
h=
f+n\lambda\sum_{i=1}^{m_0}\log\sigma(\xi_i).
\end{equation}

\begin{proposition}
Assume that $\{\xi^*_0,\cdots,\xi^*_{m_0}\}$ achieves the maximum of $h(\xi_0,\xi_1,\cdots,\xi_{m_0})$. Then
with $\lambda$ increasing, $\sigma(-\xi_i^*)(\xi_0^*-\xi_i^*)$ decreases, and, when $\lambda\to\infty$, $\sigma(-\xi_i^*)(\xi_0^*-\xi_i^*)\to0$, $\forall i=1,\cdots,m_0$.
\label{pro:3}
\end{proposition}

See Appendix C for the detailed proof. The gradient exploding issue in this local region can also be alleviated by considering fake as real. Theoretically, when the weight of fake-as-real term tends to infinity, the gradient norm of generator here becomes $0$, completely solving the concerned issue while making discriminator lose the capability of distinguishing among samples in this local region. In practice, it is enough to alleviate the gradient here to make it comparable to other gradients in a minibatch, hence we needn't weigh fake-as-real term excessively.

{\bf Alleviation for vicious circle.} Recall the vicious circle caused by gradient exploding. When more close pairs appear at an overfitting source, the fake-as-real term also turns larger from Eqn.\ref{eq:9}, providing an alleviation for a further gradient exploding issue. See the results with fake-as-real consideration applied in Fig.~\ref{fig:toy1}d, \ref{fig:toy1}h. A faithful distribution is generated even for a long time training.

\subsection{Implementation}

In this section, we give the specific implementation of Fake-As-Real GAN (FARGAN) based on gradient penalty in practical training.

For the original $N$ real samples and $M$ fake samples in a minibatch during the discriminator training process, we fix the overall number $N$ of real samples including original $N_1$ real samples and $N_0$ fake samples considered as real ones, where $N=N_0+N_1$. Note that we hope the fake samples considered as real should be in the regions where multiple close pairs exist,  because fake samples should no longer be moved towards these regions and the gradient exploding issue is relatively serious here owning to the vicious circle. For that discriminator tends to have a lower output for the region where more close pairs exist\footnote{See the proof for Proposition~\ref{pro:2} that with $m_0$ increasing, $\xi_i^*$ decreases.}, we pick out the needed $N_0$ fake samples $\widetilde{y}_{i}$ denoted as set $D_{\text{FAR}}$ as real from a larger generated set containing $f*N_0$ fake samples according to the corresponding discriminator output:
\begin{eqnarray}
D_{\text{FAR}}\!\!\!&=&\!\!\! \{\widetilde{y}_1,\cdots,\widetilde{y}_{N_0}\}=\{{y}_i, i \in \text{index of top } N_0 \text{ in}
\nonumber \\
&&\!\!\!\!\!\!\!\!\!\!\!\!\!\!\!\!\!\! \{\!-\!D_0({y}_{M\!+\!1}),\!-\!D_0({y}_{M\!+\!2}), \!\cdots\!, \!-\!D_0({y}_{M\!+\!f*N_0})\}\}.
\end{eqnarray}
When more close pairs exist, the probability of fake samples being selected in this region is higher for a lower discriminator output, in which case practical implementation still provides an alleviation for the vicious circle issue. We also add a zero-centered gradient penalty on real samples \cite{mescheder2018training} based on the discussions in Section~\ref{sec:2}, resulting in the following empirical discriminator objective in our FARGAN:
\begin{eqnarray}
\mathcal{L}_{\text{FAR}}\!\!\!\!\!\!&=&\!\!\!\!\!\! \frac{1}{N}[\sum_{i=1}^{N_1}\log(\sigma (D_0({x}_i)))+\sum_{i=1}^{N_0}\log(\sigma (D_0(\widetilde{y}_i))]
\nonumber \\
&+&\!\!\!\!\!\!\frac{1}{M}\!\sum_{i=1}^{M}\log(1\!\!-\!\sigma(D_0({y}_i))
\!\!+\!\!\frac{k}{N}\!\sum_{i=1}^{N}\!||(\nabla D_0)_{{c}_i}\!||^2\!,
\nonumber \\
\end{eqnarray}
where ${x}_i\in D_r$,${y}_i \in D_g$,$\widetilde{y}_i \in D_{\text{FAR}}$ and $\{c_1,\cdots,c_N\}=\{x_1,\cdots,x_{N_1},\widetilde{y}_1,\cdots,\widetilde{y}_{N_0}\}$.
To prevent gradient vanishing for G especially early in learning, we use the non-saturating form in the original GAN for G update. The training procedure is formally presented in Algorithm \ref{alg:1}.

\begin{algorithm}[tb]
\caption{Minibatch stochastic gradient descent training of FARGAN}
\label{alg:1}
\begin{algorithmic} 
\FOR{number of training iterations}
\WHILE{discriminator updating}
\STATE
$\bullet$ Sample minibatch of $N_1$ real examples $\{x_{1},\cdots,x_{N_1}\}$ from training dataset $D_r$.\\
$\bullet$ Sample minibatch of $M+f*N_0$ fake examples $\{y_{1},\cdots,y_{M+f*N_0}\}$ from generated dataset $D_g$.\\
$\bullet$ Determine $\widetilde{y}_{i}$ with a lower discriminator output:$\{{y}_{i}, i \in$ index of top $N_0$  in $\{-D_0({y}_{M+1})$, $\cdots$,$-D_0({y}_{M+f*N_0})\}\}$.\\
$\bullet$ Update the discriminator by ascending its stochastic gradient: $\nabla_{\theta_d}\mathcal{L}_{\text{FAR}}$.\\
\ENDWHILE
\STATE
$\bullet$ Sample minibatch of $M$ fake examples $\{y_{1},\cdots,y_{M}\}$ from generated dataset $D_g$.\\
$\bullet$ Update the generator by ascending its stochastic gradient: $\nabla_{\theta_g}\frac{1}{M}\sum_{i=1}^{M}\log(\sigma (D_0({y}_i)))$.
\ENDFOR
\end{algorithmic}
\end{algorithm}

\section{Experiments}

In this section, we present our experimental results on synthetic data and real-world datasets including CIFAR-10 \cite{Antonio200880}, CIFAR-100 \cite{Antonio200880} and a more challenging dataset ImageNet \cite{russakovsky2015imagenet}. When we talk the fake-as-real method, a zero-centered gradient penalty on real samples is also added as a default in our experiments.  We use Pytorch \cite{paszke2017automatic} for development.

\subsection{Synthetic data}
To test the effectiveness of FARGAN on preventing an unbalanced generation, we designed a dataset with finite training samples coming from a Gaussian distribution. Based on a simple MLP network, we trained Non-Saturating GAN (NSGAN) with our method and different gradient penalties including zero-centered gradient penalty on real samples (NSGAN-0GP-sample) and on interpolation between real and fake samples (NSGAN-0GP-interpolation). We set the weight $k$ of gradient penalty to be $10$, the size of minibatch $N=M=64$ and $f=8$, $N_0=16$ for FARGAN. Learning rate is set to be $0.003$ for both G and D. The result is shown in Fig.~\ref{fig:toy1}. It can be observed that NSGAN, NSGAN-0GP-sample and NSGAN-0GP-interpolation all generate unbalanced distributions as training progresses, while our method can generate much better results with good generalization.

\begin{figure}[h]
  \centering
  \subfigure[]{\includegraphics[width=1.3in]{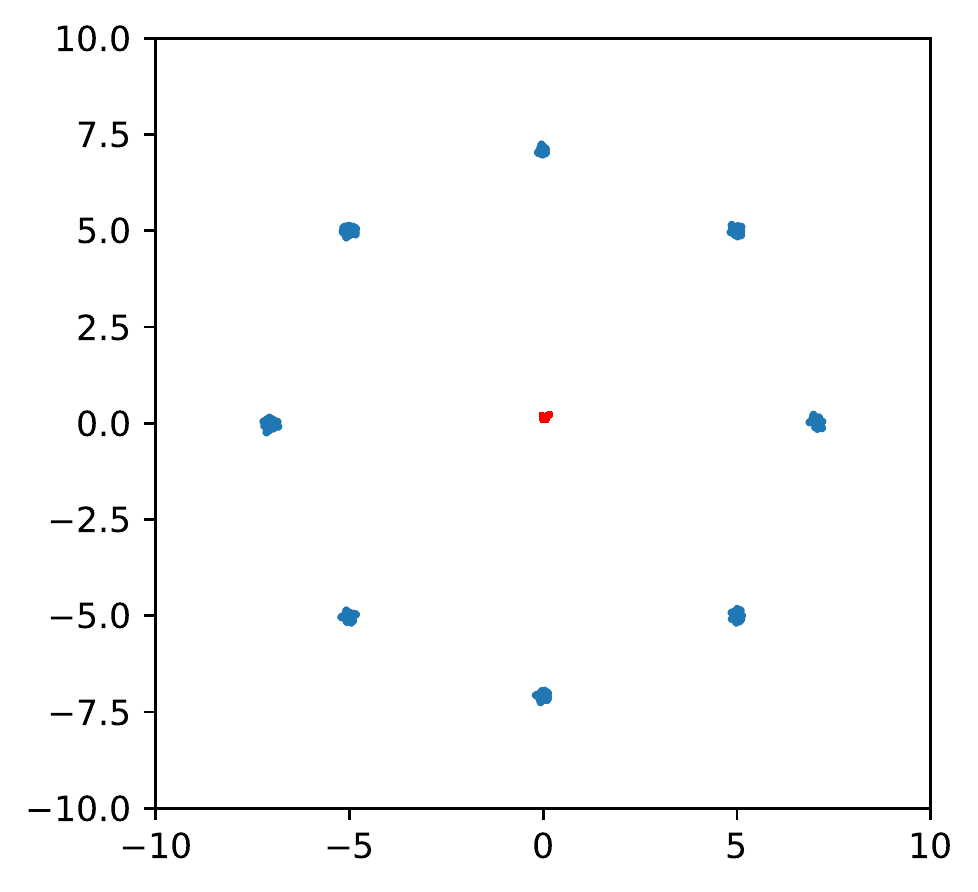}}
    \centering
  \subfigure[]{\includegraphics[width=1.3in]{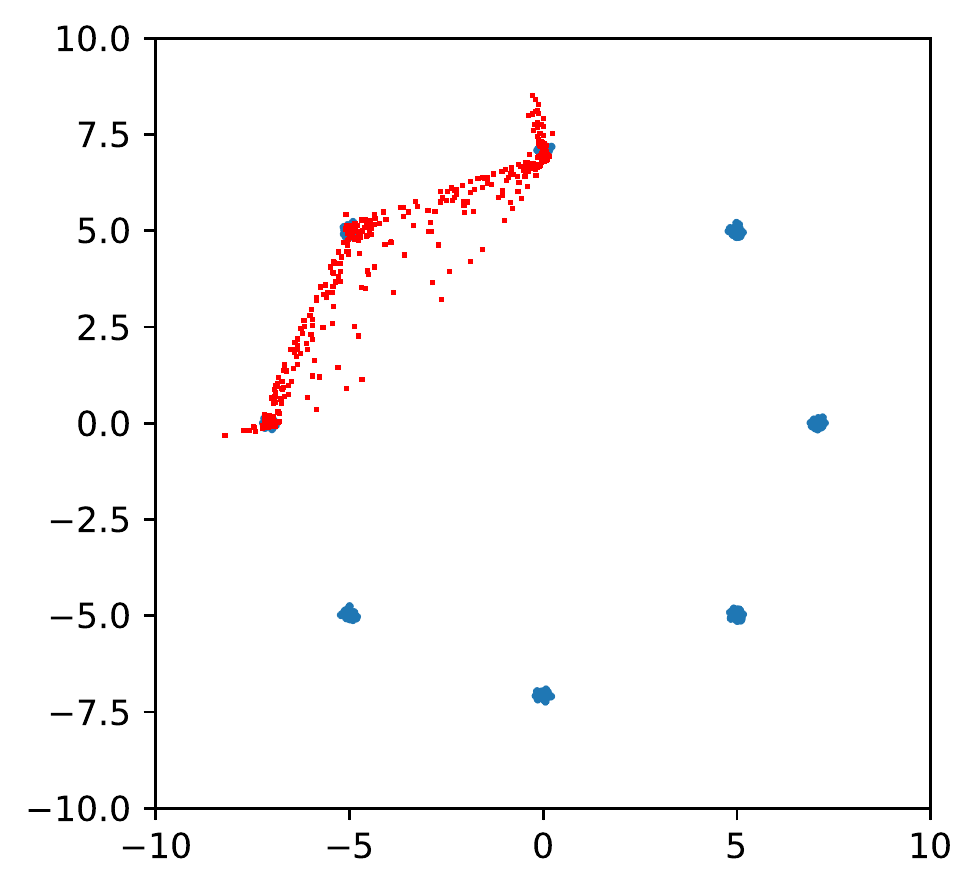}}
   \centering
  \subfigure[]{\includegraphics[width=1.3in]{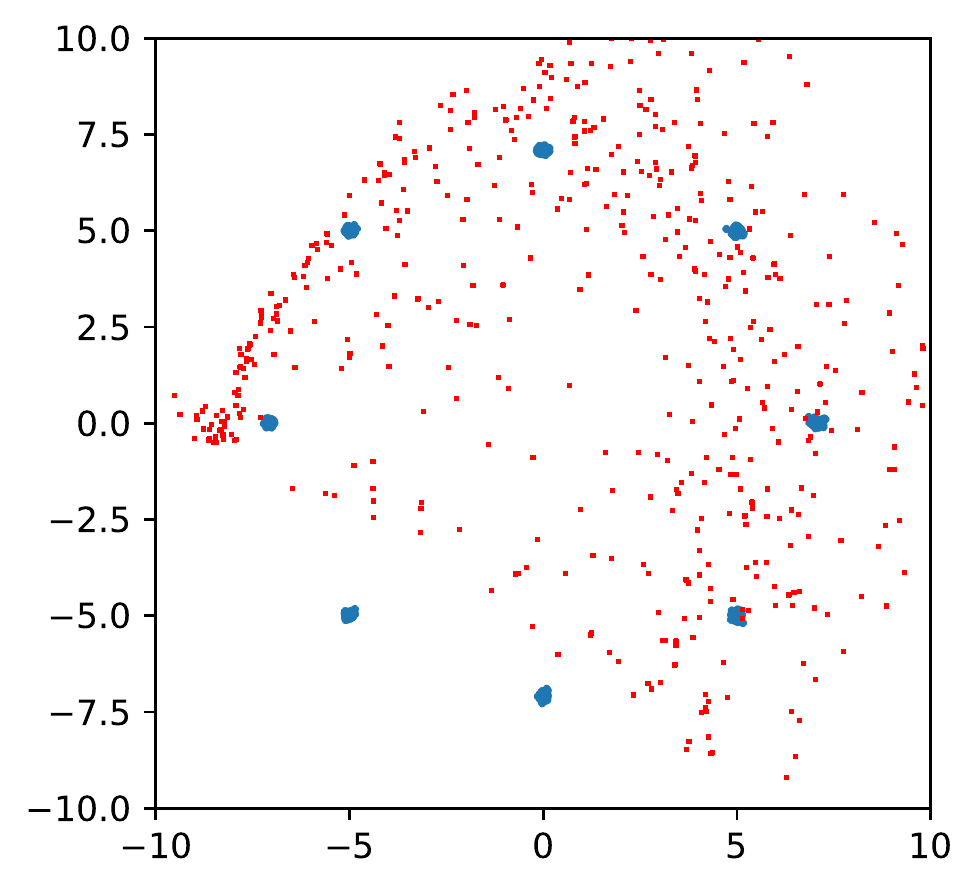}}
   \centering
  \subfigure[]{\includegraphics[width=1.3in]{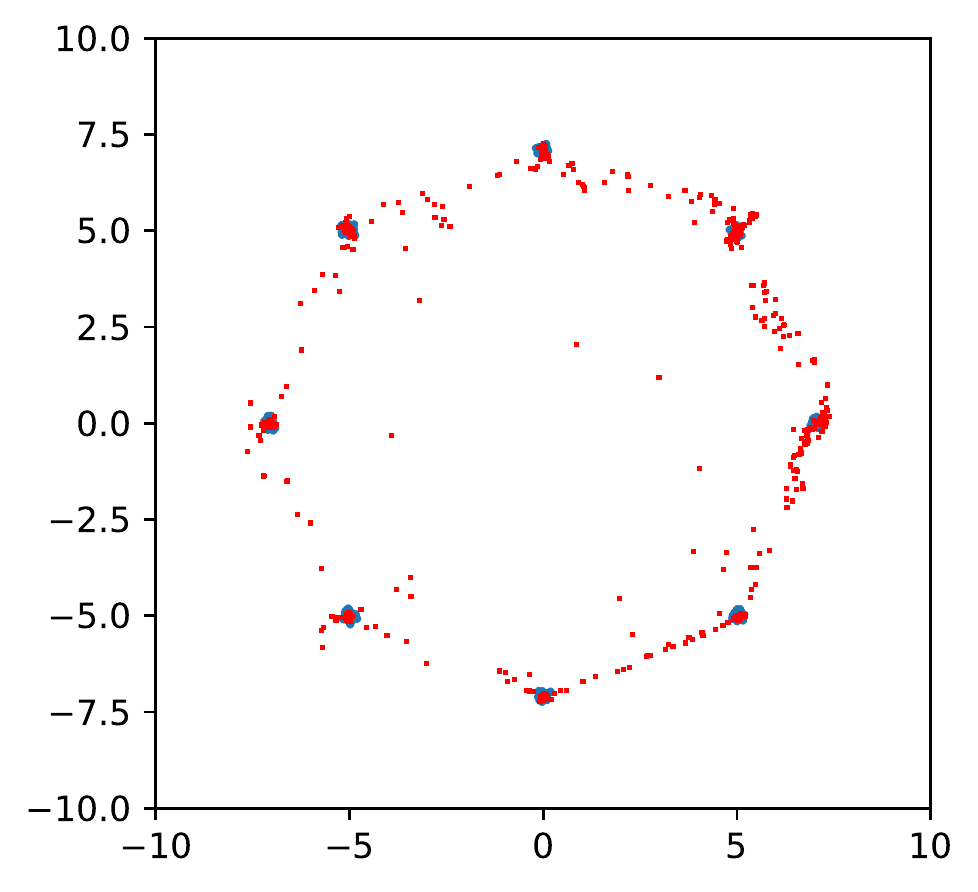}}
  \caption{Evolution of our method on a mixture of $8$ Gaussians dataset. (a) iter. 0. (b) iter. 100k. (c) iter. 335k. (d) iter. 500k. }
  \label{fig:toy2}
\end{figure}

We also test FARGAN on a mixture of $8$ Gaussians dataset where random samples in different modes are far from each other. The evolution of FARGAN is depicted in Fig.\ref{fig:toy2}. Although FARGAN only covers $3$ modes at the beginning, it can cover other modes gradually for the powerful capability of gradient exploding alleviation. Hence, FARGAN has the ability to find the uncovered modes to achieve a faithful distribution even when samples in high dimensional space are far from each other. More synthetic experiments can be found in Appendix E.

\begin{table*}
\centering
\caption{Inception score and FID on CIAFR-10, CIFAR-100 at iter. 500k and ImageNet at iter. 600k. Experiments were repeated 3 times.}
  \begin{tabular}{ c  c  c  c  c }
    \hline
  \multirow{2}{*}{} &
  \multicolumn{2}{c}{IS}        &
  \multicolumn{2}{c}{FID}        \\
    \cline{2-5} &
  0GP         & FAR   &
  0GP        & FAR     \\
    \hline
  CIFAR-10 (500k) & &        &  &
             \\
  ResNet NSGAN &$6.26\pm 0.09$ & $6.81\pm 0.03$      & $24.22\pm 0.72$ & $17.82\pm 0.33$
             \\
  ResNet WGAN & $6.15\pm 0.06$& $6.83\pm 0.04$       & $24.72\pm 0.41$ & $18.12\pm 0.23$
             \\
  ResNet HingeGAN &$6.19\pm 0.08$ & $6.88\pm 0.07$      & $24.55\pm 0.31$ & $16.99\pm 0.18$
             \\
  ResNet LSGAN & $5.90\pm 0.05$ & $6.63\pm 0.02$       & $26.41\pm 0.12$ & $19.97\pm 0.38$
             \\
  Conventional NSGAN & $6.94\pm 0.03$ & $7.63\pm 0.05$       & $16.66\pm 0.14$ & $12.80\pm 0.31$
             \\
  \hline
  CIFAR-100 (500k) &  &       &  &
             \\
  ResNet NSGAN & $6.27\pm 0.04$ & $7.03\pm 0.06$       & $28.46\pm 0.28$ & $21.95\pm 0.35$
             \\
  Conventional NSGAN & $6.92\pm 0.08$ & $7.84\pm 0.04$       & $22.28\pm 0.45$ & $17.69\pm 0.24$
             \\
  \hline
  ImageNet (600k) &  &       & &
             \\
  ResNet NSGAN & $10.66\pm 0.11$ & $11.44\pm 0.05$       & $44.57\pm 0.34$ & $39.69\pm 0.57$
             \\
  \hline
  \end{tabular}
  \label{tab:1}
\end{table*}

\subsection{CIFAR-10 and CIFAR-100}
In this section, we compare the fake-as-real method with that with only zero-centered gradient penalty (0GP) on real samples added. All experiments are repeated $3$ times with random initialization to show the consistent results in Tab.~\ref{tab:1}.

{\bf Parameter settings.}
We set the weight $k$ of gradient penalty to be $10$, the size of minibatch $N=M=64$ and $f=8$, $N_0=32$ for fake-as-real method as a default.
RMSProp optimizer with $\alpha=0.99$ and a learning rate of $10^{-4}$ is used.

{\bf Quantitative measures.}
Inception score \cite{SalimansGZCRCC16} and FID \cite{heusel2017gans} are used as quantitative measures. For Inception score, we follow the guideline from \cite{SalimansGZCRCC16}. The FID score is evaluated on $10k$ generated images. Better generation can be achieved with higher inception score and lower FID value.

\begin{figure}
  \centering
  \subfigure[]{\includegraphics[width=2.6in,height=1.6in]{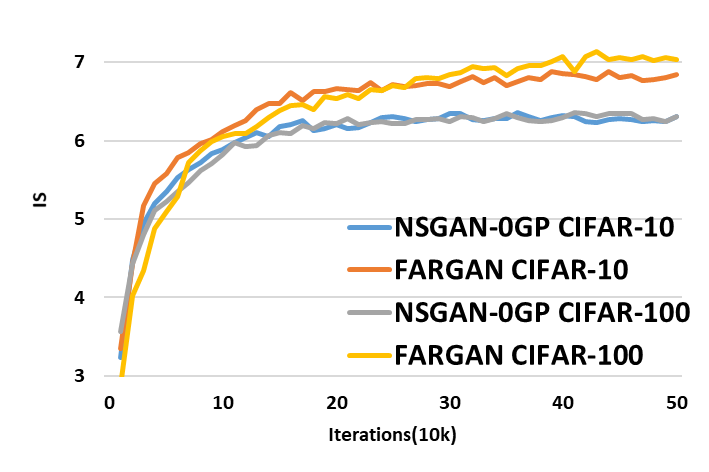}}
    \centering
  \subfigure[]{\includegraphics[width=2.6in,height=1.6in]{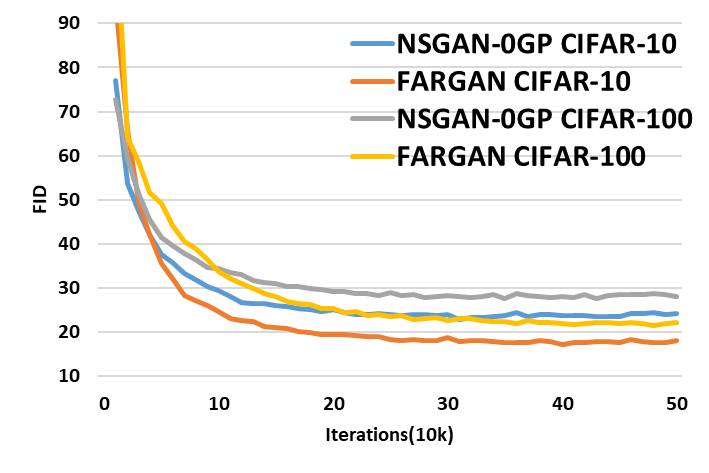}}
  \caption{Results with ResNet architecture on CIFAR dataset.}
  \label{fig:Arc_ResNet}
\end{figure}

\begin{figure}
  \centering
  \subfigure[]{\includegraphics[width=2.6in,height=1.6in]{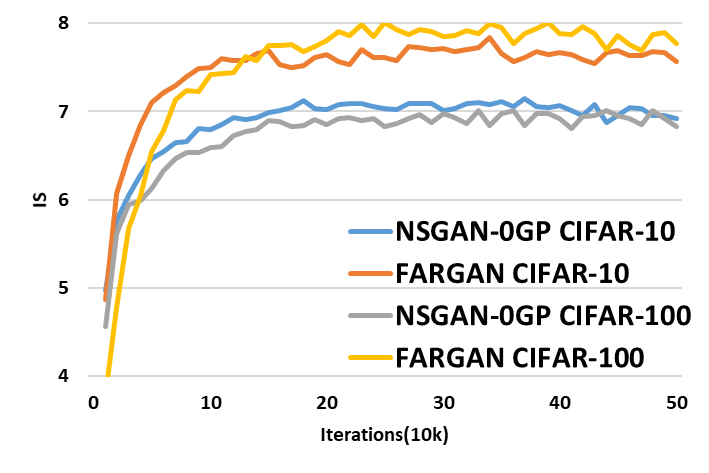}}
    \centering
  \subfigure[]{\includegraphics[width=2.6in,height=1.6in]{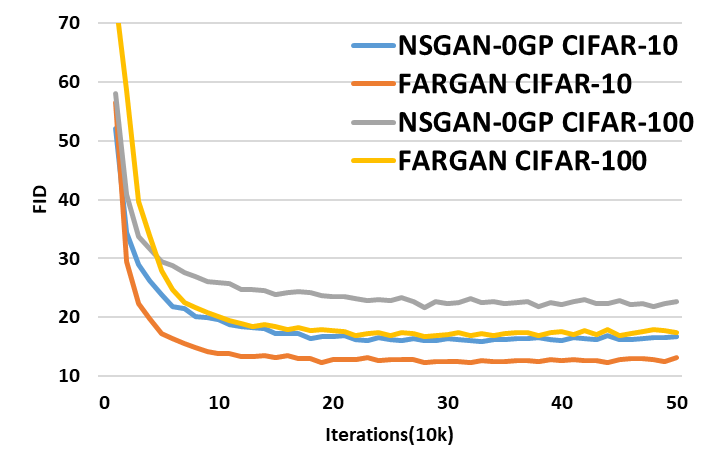}}
  \caption{Results with conventional architecture on CIFAR dataset.}
  \label{fig:Arc_convention}
  \vspace{-0.2cm}
\end{figure}

\begin{figure}[h]
  \centering
  \subfigure[]{\includegraphics[width=2.6in,height=1.6in]{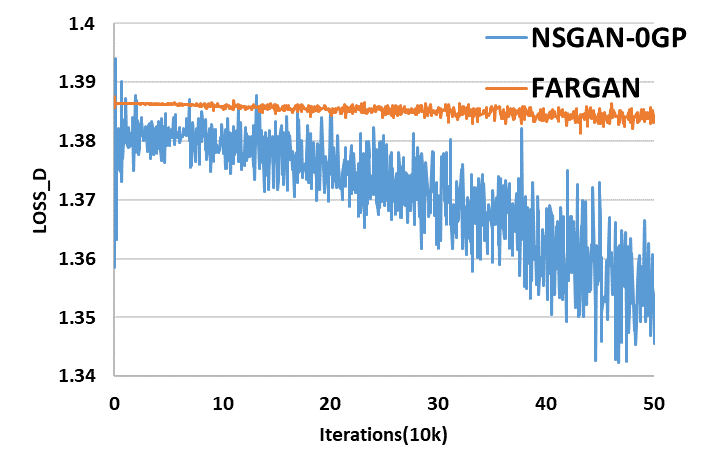}}
    \centering
  \subfigure[]{\includegraphics[width=2.6in,height=1.6in]{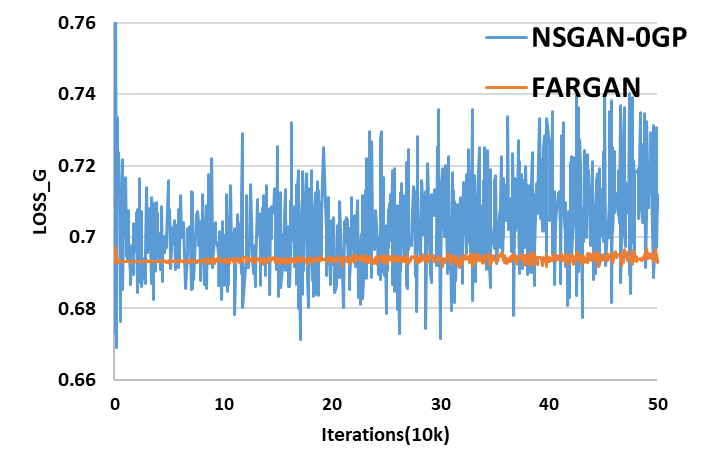}}
  \caption{Losses of discriminator (not including regularization term) and generator on CIFAR-10.}
  \label{fig:Stability}
  \vspace{-0.2cm}
\end{figure}

\begin{figure}
  \centering
  \subfigure[]{\includegraphics[width=2.6in,height=1.6in]{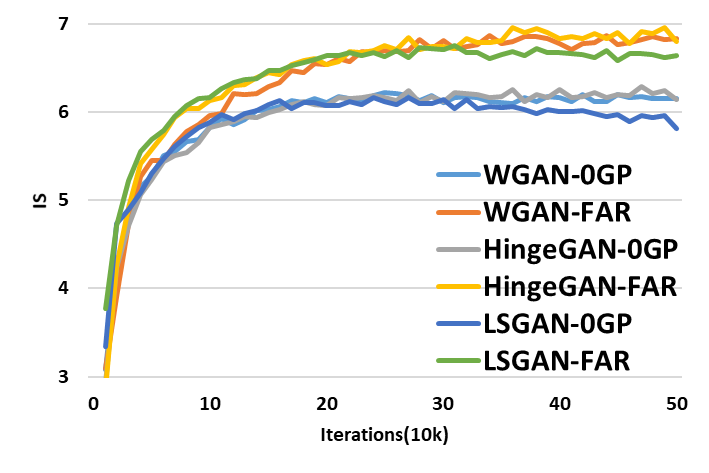}}
    \centering
  \subfigure[]{\includegraphics[width=2.6in,height=1.6in]{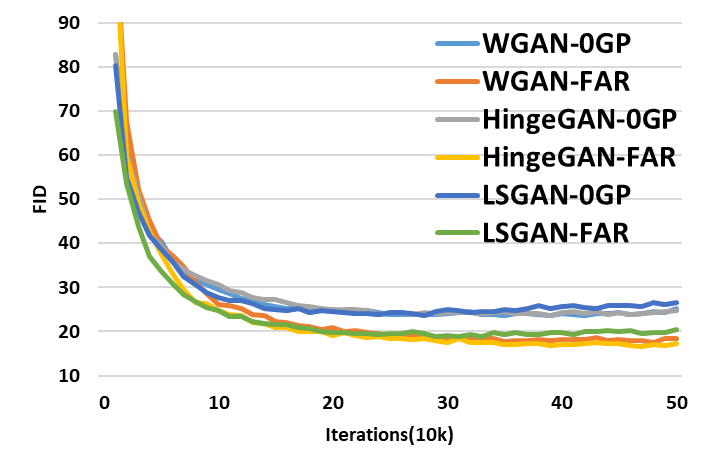}}
  \caption{Results of different GAN variants on CIFAR-10. }
  \label{fig:Variants}
  \vspace{-0.2cm}
\end{figure}

{\bf Results with different architectures.}
We test FARGAN with both a ResNet architecture the same as that in \cite{mescheder2018training} and a conventional architecture similar to a progressively growing GAN \cite{karras2018progressive} while with no batch normalization. The results are shown in Fig.~\ref{fig:Arc_ResNet} and \ref{fig:Arc_convention} respectively. FARGAN outperforms NSGAN-0GP with both architectures on CIFAR-10 and CIFAR-100 by a large margin. Note although the speed of FARGAN to cover real
ones could be slightly slowed down at the beginning of training with some fake samples considered as real ones, it can consistently improve the results of generation and achieve a more balanced distribution finally.

The losses of discriminator and generator during the training process with ResNet architecture on CIFAR-10 are shown in Fig.\ref{fig:Stability}. FARGAN has a much more stable training process with smaller fluctuations and no obvious deviation seen for the losses. Note when serious mode collapse happens, discriminator has a lower loss while generator has a higher loss compared with the theoretical value ($2\log2\approx 1.386$ for discriminator and $\log2\approx 0.693$ for generator)\footnote{Discriminator outputs a high value for uncovered modes while a low value for over-covered modes.}. The gradual deviation of losses for discriminator and generator in NSGAN-0GP shows a serious mode collapse. Hence, FARGAN can stabilize training process and effectively prevent mode collapse. The losses of discriminator and generator on CIFAR-100 and generated image samples can be found in Appendix E.

{\bf Results of different GAN-variants.}
Besides NSGAN, we also test fake-as-real method for WGAN \cite{pmlr-v70-arjovsky17a}, HingeGAN \cite{zhao2016energy} and LSGAN \cite{mao2017least} to show the effectiveness on a more faithful generation for different GAN-variants.  The results are shown in Fig.~\ref{fig:Variants}. Fake-as-real method can also improve the performance of different GAN-variants by alleviating the gradient exploding issue which consistently happens for finite training samples.

\begin{figure}
  \centering
  \subfigure[]{\includegraphics[width=2.6in,height=1.6in]{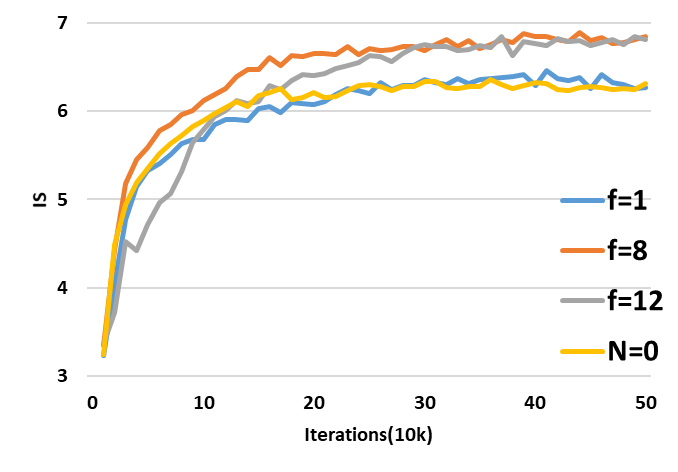}}
    \centering
  \subfigure[]{\includegraphics[width=2.6in,height=1.6in]{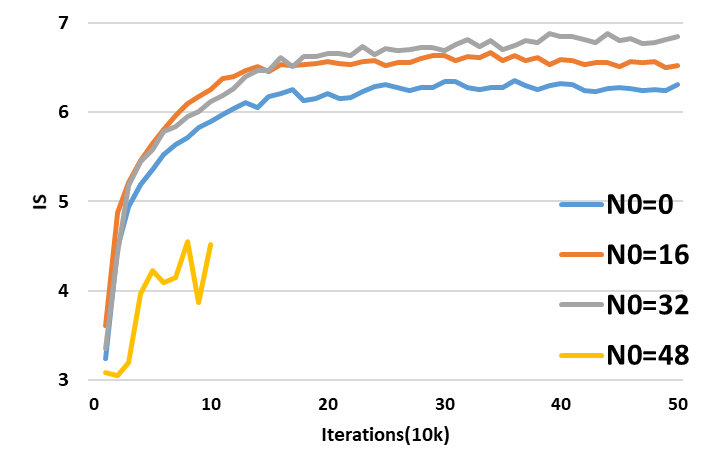}}
  \caption{Results of FARGAN with different $f$ and $N_0$.}
  \label{fig:f_N0}
  \vspace{-0.3cm}
\end{figure}

{\bf Results with different $f$ and $N_0$ in FARGAN.} We make an ablation study on the selection of parameters $f$ and $N_0$ in FARGAN. With ResNet architecture on CIFAR-10, we first fix $N_0=32$ and change the value of $f$. Then we fix $f=8$ and change the value of $N_0$. The results are shown in Fig.~\ref{fig:f_N0}. Note that the training speed could be slightly slowed down with $f$ and $N_0$ increasing while a better generation could be achieved. An obvious improvement is achieved with $f$ increasing until $f$ is big enough, e.g. $f=8$. An improvement is also seen with $N_0$ increasing appropriately while a collapse happens when $N_0$ is too big e.g. $N_0=48$, for the too weak capability of discriminator. Hence, in practice we set  $f=8$ and $N_0=32$ as a default.

Note that when $f=1$, we select fake samples randomly as real ones, and, when $N_0=0$, no fake samples are considered as real ones. We observe that an obvious improvement is not achieved for FARGAN with $f=1$ compared with $N_0=0$. However, FARGAN with $f=8$ improves the performance by a large margin. Hence, the key point is considering fake samples in the gradient exploding regions instead of selected randomly as real ones according to our theoretical analysis and experiments.

\subsection{ImageNet}

\begin{figure}
  \centering
  \subfigure[]{\includegraphics[width=2.6in,height=1.6in]{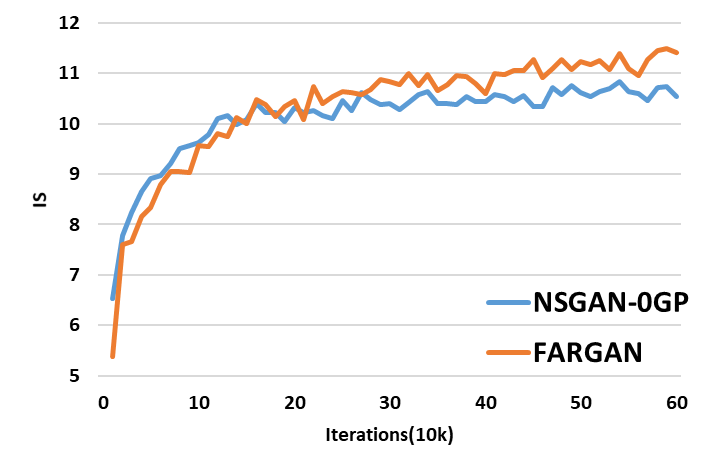}}
    \centering
  \subfigure[]{\includegraphics[width=2.6in,height=1.6in]{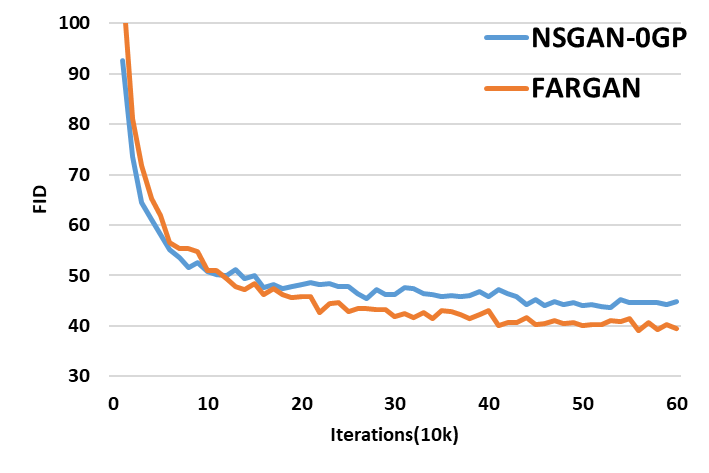}}
  \caption{Results on ImageNet. }
  \label{fig:ImageNet}
  \vspace{-0.3cm}
\end{figure}

For the challenging ImageNet task which contains $1000$ classes, we train GANs with ResNet architecture to learn generative models. We use images at resolution $64\times64$ and no labels are used in our models. We use the Adam optimizer with $\alpha=0$, $\beta=0.9$. Other settings are the same as that in CIFAR experiments.
The results in Fig.\ref{fig:ImageNet} show that FARGAN still outperforms NSGAN-0GP on ImageNet and produces samples of state of the art quality without using any labels or particular architectures like progressive
growing trick \cite{karras2018progressive}.  Random selected samples and losses of discriminator and generator during the training process can be found in Appendix E.

\section{Conclusion}

In this paper, we explain the reason that an unbalanced distribution is often generated in GANs. We show that the existence of vicious circle resulted from gradient exploding, makes unbalanced generation more and more serious as training progresses. We analyze methods of gradient exploding alleviation including  difference penalization between discriminator outputs on close real and fake pairs and trick of considering fake as real. Based on the theoretical analysis, we propose FARGAN by considering fake as real according to the discriminator
outputs in a training minibatch. Experiments on diverse datasets verify that our method can stabilize the
training process and improve the performance by a large margin.

\section*{Acknowledgement}
This work was supported by the National Natural Science Foundation of China under grant (61771305, 61771303), and Science and Technology Commission of Shanghai Municipality (STCSM, Grant No.18DZ1200102).

{\small
\bibliographystyle{ieee_fullname}
\bibliography{egbib}
}

\appendix
\section{Proof for Proposition 1}
\label{app1}
For empirical discriminator, it maximizes the following objective:
\begin{equation}
\mathcal{L}= \mathbb{E}_{{x}\in D_r}[\log(D({x}))]+\mathbb{E}_{{y}\in D_g}[\log(1-D({y}))].
\end{equation}
Assume that samples are normalized:
\begin{equation}
||{x_i}||=||{y_i}||=1, \forall {x} \in D_r,{y} \in D_g.
\end{equation}
Let ${W_1}\in \mathbb{R}^{2\times d_x}$, ${W_2}\in \mathbb{R}^{2\times2}$ and ${W_3}\in \mathbb{R}^{2}$ be the weight matrices, ${b}\in \mathbb{R}^{2}$ offset vector and $k_1$,$k_2$ a constant, We can construct needed discriminator as a MLP with two hidden layer containing $\mathcal{O}(2dim(x))$ parameters. We set weight matrices
\begin{equation}
{W_1}=\left[
 \begin{matrix}
   {x_0}^T  \\
   {y_0}^T  \\
  \end{matrix}
  \right],
  {W_2}=\left[
 \begin{matrix}
   1  & -1  \\
   -1 & 1  \\
  \end{matrix}
  \right],
 {W_3}=\left[
 \begin{matrix}
   \frac{1}{2}+\frac{\epsilon}{2}  \\
   \frac{1}{2}-\frac{\epsilon}{2}  \\
  \end{matrix}
  \right].
\end{equation}
For any input ${v}\in D_r \cup D_g$, the discriminator output is computed as:
\begin{equation}
D({v})={W_3}^T\sigma (k_2{W_2}\sigma(k_1({W_1}{v}-{b}))),
\end{equation}
where $\sigma(x)=\frac{1}{1+e^{-x}}$ is the sigmoid function.
Let $\alpha={W_1}{v}-{b}$, we have
\begin{equation}
\alpha_1=\left\{
\begin{aligned}
1-b_1,\text{if } {v}={x_0} \\
l-b_1,\text{if } {v}\neq {x_0} \\
\end{aligned}
\right.,
\alpha_2=\left\{
\begin{aligned}
1-b_2,\text{if } {v}={y_0} \\
l-b_2,\text{if } {v}\neq {y_0} \\
\end{aligned}
\right.,
\end{equation}
where $l<1$. Let $\beta=\sigma(k_1\alpha)$, we have
\begin{equation}
\beta_1=\left\{
\begin{aligned}
1,\text{if } {v}={x_0} \\
0,\text{if } {v}\neq {x_0} \\
\end{aligned}
\right.,
\beta_2=\left\{
\begin{aligned}
1,\text{if } {v}={y_0} \\
0,\text{if } {v}\neq {y_0} \\
\end{aligned}
\right.
\end{equation}
as $k_1\rightarrow \infty$ and $b\rightarrow 1^-$. Let $\gamma=\sigma(k_2{W_2}\beta)$, we have
\begin{equation}
\gamma_1=\left\{
\begin{aligned}
1&,&\text{if } {v}={x_0} \\
0&,&\text{if } {v}={y_0} \\
\frac{1}{2}&,&\text{if } {v}\neq {x_0},{y_0} \\
\end{aligned}
\right.,
\gamma_2=\left\{
\begin{aligned}
0&,&\text{if } {v}={x_0} \\
1&,&\text{if } {v}={y_0} \\
\frac{1}{2}&,&\text{if } {v}\neq {x_0},{y_0} \\
\end{aligned}
\right.
\end{equation}
as $k_2\rightarrow \infty$.
Hence, for any input ${v}\in D_r \cup D_g$, discriminator outputs
\begin{equation}
D({v})={W_3}^T\gamma=\left\{
\begin{aligned}
\frac{1}{2}+\frac{\epsilon}{2},&\text{if } {v}={x_0} \\
\frac{1}{2}-\frac{\epsilon}{2},&\text{if } {v}={y_0} \\
\frac{1}{2}~~~~,&\text{else }  \\
\end{aligned}
\right..
\end{equation}
In this case, the discriminator objective has a more optimal value than the theoretical optimal version:
\begin{eqnarray}
\mathcal{L}&=&\frac{1}{n}((n-1)\log\frac{1}{2}+\log(\frac{1}{2}+\frac{\epsilon}{2}))
\nonumber \\
&&+
\frac{1}{m}((m-1)\log\frac{1}{2}+\log(\frac{1}{2}+\frac{\epsilon}{2}))
\nonumber\\
&>&2\log\frac{1}{2}.
\end{eqnarray}
Then the discriminator outputs a constant $\frac{1}{2}$ except that $D(x_0)=\frac{1}{2}+\frac{\epsilon}{2}$ and $D(y_0)=\frac{1}{2}-\frac{\epsilon}{2}$ satisfying the condition.

\section{Proof for proposition 2}
\label{app2}
We rewrite $f(\xi_0,\xi_1,\cdots,\xi_{m_0})$ here
\begin{equation}
f\!=\!\log\sigma(\xi_0)\!+\!\frac{n}{m}\sum_{i=1}^{m_0}\log(1\!-\!\sigma(\xi_i))\!-\!\frac{nk}{m_0}\sum_{i=1}^{m_0}(\xi_0\!-\!\xi_i)^2.
\end{equation}
To achieve the optimal value, let $f'(\xi_i)=0,i=0,\cdots,m_0$ and we have
\begin{eqnarray}
f'(\xi^*_0)\!\!\!\!\!&=&\!\!\!\!\!
1-\sigma(\xi^*_0)-\frac{2nk}{m_0}\sum_{i=1}^{m_0}(\xi^*_0-\xi^*_i)=0,
\\
f'(\xi^*_i)\!\!\!\!\!&=&\!\!\!\!\!\!
-\!\frac{n}{m}\sigma(\xi^*_i)\!+\!\frac{2nk}{m_0}(\xi^*_0\!-\!\xi^*_i)\!=\!0
,i\!=\!1,\!\cdots\!,m_0.
\nonumber \\
\end{eqnarray}
It is obvious that $\xi^*_1=\xi^*_2=\cdots=\xi^*_{m_0}=\xi^*$. Hence we have
\begin{eqnarray}
1-\sigma(\xi^*_0)-2nk(\xi^*_0-\xi^*)=0,
\label{eq:a4}
\\
-\frac{n}{m}\sigma(\xi^*)+\frac{2nk}{m_0}(\xi^*_0-\xi^*)=0.
\end{eqnarray}
We can solve
\begin{equation}
\xi^*=
-\ln(\frac{nm_0}{m\sigma(-\xi^*_0)}-1).
\label{eq:a3}
\end{equation}
Substitute Eqn. \ref{eq:a3} into Eqn. \ref{eq:a4} and we get
\begin{equation}
f'(\xi^*_0)=\sigma(-\xi^*_0)-2nk(\xi^*_0+\ln(\frac{nm_0}{m\sigma(-\xi^*_0)}-1))=0.
\label{eq:a5}
\end{equation}
We can also have from Eqn. \ref{eq:a3} and Eqn. \ref{eq:a4} respectively
\begin{eqnarray}
\xi^*_0-\xi^*&=&\xi^*_0+\ln(\frac{nm_0}{m\sigma(-\xi^*_0)}-1),
\label{eq:a6}
\\
&=&\frac{\sigma(-\xi^*_0)}{2nk}.
\label{eq:a7}
\end{eqnarray}
To satisfy Eqn.\ref{eq:a5}, $\xi_0^*+\log(\frac{nm_0}{m\sigma(-\xi_0^*)}-1)>0$. Hence, with $k$ increasing, $\xi_0^*$ decreases from Eqn.\ref{eq:a5}. Based on Eqn.\ref{eq:a6}, we further know with $k$ increasing, $\xi^*$ increases, $\xi_0^*-\xi^*$ decreases and $\sigma(-\xi_i^*)(\xi_0^*-\xi_i^*)$ decreases. Similarly, based on Eqn.\ref{eq:a7} and Eqn.\ref{eq:a5}, we can achieve with $m_0$ increasing, $\sigma(-\xi_i^*)(\xi_0^*-\xi_i^*)$ increases finishing the proof.

\section{Proof for proposition 3}
\label{app3}
Similar to the proof for Proposition 2, let $h'(\xi_0)=h'(\xi_i)=0,i=1,\cdots,m_0$, and we can easily achieve $\xi_1^*=\xi_2^*=\cdots=\xi_{m_0}^*$,
\begin{equation}
\lambda(\xi_0^*)=\frac{\sigma(-\xi_0^*)}{nm_0}
[e^{2nk\xi_0^*-\sigma(-\xi_0^*)}
(\frac{nm_0}{m\sigma(-\xi_0^*)}-1)^{2nk}-1],
\label{eq:a8}
\end{equation}
and
\begin{equation}
\xi_0^*-\xi_i^*=\frac{\sigma(-\xi_0^*)}{2nk}.
\label{eq:a9}
\end{equation}
It can be easily proved that $\lambda'(\xi_0^*)>0$.
To satisfy Eqn.\ref{eq:a8}, with $\lambda$ increasing, $\xi_0^*$ increases, and, when $\lambda\to\infty$, $\xi_0^*\to\infty$.
Based on Eqn.\ref{eq:a9}, we further know with $\lambda$ increasing, $\xi_i^*$ increases, $\xi_0^*-\xi_i^*$ decreases, and $\sigma(-\xi_i^*)(\xi_0^*-\xi_i^*)$ decreases. We can also achieve that when $\lambda\to\infty$, $\sigma(-\xi_i^*)(\xi_0^*-\xi_i^*)\to0$, finishing the proof.

\section{Network architectures}
\label{app4}
For synthetic experiment, the network architectures are the same as that in \cite{thanh2019improving}. While for real world data experiment, we use both the ResNet \cite{mescheder2018training} and conventional \cite{karras2018progressive} architecture and a simple DCGAN based architecture \cite{radford2015unsupervised} is also included.

\begin{table}[H]
		\begin{center}
		\caption{Generator architecture in synthetic experiment }
		\begin{tabular}{|c c c|}
		    \hline
		    Layer & output size &filter \\
		    \hline
			Fully connected & $64$ & $2 \rightarrow 64$\\
			RELU & $64$ & -\\
		    \hline
			Fully connected & $64$ & $64 \rightarrow 64$\\
			RELU & $64$ & -\\
			\hline
            Fully connected & $64$ & $64 \rightarrow 64$\\
			RELU & $64$ & -\\
			\hline
            Fully connected & $2$ & $64 \rightarrow 2$\\
			\hline
		\end{tabular}
		\end{center}
	\end{table}

\begin{table}[H]
		\begin{center}
		\caption{Discriminator architecture in synthetic experiment }
		\begin{tabular}{|c c c|}
		    \hline
		    Layer & output size &filter \\
		    \hline
			Fully connected & $64$ & $2 \rightarrow 64$\\
			RELU & $64$ & -\\
		    \hline
			Fully connected & $64$ & $64 \rightarrow 64$\\
			RELU & $64$ & -\\
			\hline
            Fully connected & $64$ & $64 \rightarrow 64$\\
			RELU & $64$ & -\\
			\hline
            Fully connected & $1$ & $64 \rightarrow 1$\\
			\hline
		\end{tabular}
		\end{center}
	\end{table}

\begin{table}[H]
		\begin{center}
		\caption{Generator DCGAN based architecture in CIFAR experiment }
		\begin{tabular}{|c c c|}
		    \hline
		    Layer & output size &filter \\
		    \hline
			Fully connected & $256 \cdot 4 \cdot 4$ & $128 \rightarrow 256 \cdot 4\cdot4$\\
			Reshape & $256\times4\times4$ & -\\
            TransposedConv2D & $128\times8\times8$ & $256 \rightarrow 128$\\
            TransposedConv2D & $64\times16\times16$ & $128 \rightarrow 64$\\
            TransposedConv2D & $3\times32\times32$ & $64 \rightarrow 3$\\
			\hline
		\end{tabular}
		\end{center}
	\end{table}
	
\begin{table}[H]
		\begin{center}
		\caption{Discriminator DCGAN based architecture in CIFAR experiment }
		\begin{tabular}{|c c c|}
		    \hline
		    Layer & output size &filter \\
			\hline
            Conv2D & $64\times16\times16$ & $3 \rightarrow 64$\\
            Conv2D & $128\times8\times8$ & $64 \rightarrow 128$\\
            Conv2D & $256\times4\times4$ & $128 \rightarrow 256$\\
			Reshape & $256\cdot4\cdot4$ & -\\
			Fully Connected & $256\cdot4\cdot4$ & $256\cdot4\cdot4\rightarrow1$\\
			\hline
		\end{tabular}
		\end{center}
	\end{table}

\begin{table}[H]
		\begin{center}
		\caption{Generator ResNet architecture in CIFAR experiment }
		\begin{tabular}{|c c c|}
		    \hline
		    Layer & output size &filter \\
		    \hline
			Fully connected & $512 \cdot 4 \cdot 4$ & $128 \rightarrow 512 \cdot 4\cdot4$\\
			Reshape & $512\times4\times4$ & -\\
			\hline
			Resnet-Block & $256\times4\times4$ & $512\rightarrow256\rightarrow256$\\
			NN-Upsampling & $256\times8\times8$ & -\\
			\hline
			Resnet-Block & $128\times8\times8$ & $256\rightarrow128\rightarrow128$\\
			NN-Upsampling & $128\times16\times16$ & -\\
			\hline
			Resnet-Block & $64\times16\times16$ & $128\rightarrow64\rightarrow64$\\
			NN-Upsampling & $64\times32\times32$ &  -\\
			\hline
			Resnet-Block & $64\times32\times32$ & $64\rightarrow64\rightarrow64$\\
			Conv2D & $3\times32\times32$ & $64\rightarrow3$\\
			\hline
		\end{tabular}
		\end{center}
	\end{table}
	
\begin{table}[H]
		\begin{center}
		\caption{Discriminator ResNet architecture in CIFAR experiment }
		\begin{tabular}{|c c c|}
		    \hline
		    Layer & output size &filter \\
		    \hline
			Conv2D & $64\times32\times32$ & $3 \rightarrow 64$\\
			\hline
			Resnet-Block & $128\times32\times32$ & $64\rightarrow64\rightarrow128$\\
			Avg-Pool2D & $128\times16\times16$ & -\\
			\hline
			Resnet-Block & $256\times16\times16$ & $128\rightarrow128\rightarrow256$\\
			Avg-Pool2D & $256\times8\times8$ & -\\
			\hline
			Resnet-Block & $512\times8\times8$ & $256\rightarrow256\rightarrow512$\\
			Avg-Pool2D & $512\times4\times4$ &  -\\
			\hline
			Reshape & $512\cdot4\cdot4$ & -\\
			Fully Connected & $1$ & $512\cdot4\cdot4\rightarrow1$\\
			\hline
		\end{tabular}
		\end{center}
	\end{table}

\begin{table}[H]
		\begin{center}
		\caption{Generator conventional architecture in CIFAR experiment }
		\begin{tabular}{|c c c|}
		    \hline
		    Layer & output size &filter \\
		    \hline
			Fully connected & $512 \cdot 1 \cdot 1$ & $128 \rightarrow 512 \cdot 1\cdot1$\\
			Reshape & $512\times1\times1$ & -\\
			\hline
            TransposedConv2D & $512\times4\times4$ & $512 \rightarrow 512$\\
            TransposedConv2D & $512\times4\times4$ & $512 \rightarrow 512$\\
            NN-Upsampling & $512\times8\times8$ & -\\
            \hline
            TransposedConv2D & $256\times8\times8$ & $512 \rightarrow 256$\\
            TransposedConv2D & $256\times8\times8$ & $256 \rightarrow 256$\\
            NN-Upsampling & $256\times16\times16$ & -\\
            \hline
            TransposedConv2D & $128\times16\times16$ & $256 \rightarrow 128$\\
            TransposedConv2D & $128\times16\times16$ & $128 \rightarrow 128$\\
            NN-Upsampling & $128\times32\times32$ & -\\
            \hline
            TransposedConv2D & $64\times32\times32$ & $128 \rightarrow 64$\\
            TransposedConv2D & $64\times32\times32$ & $64 \rightarrow 64$\\
            TransposedConv2D & $3\times32\times32$ & $64 \rightarrow 3$\\
			\hline
		\end{tabular}
		\end{center}
	\end{table}
	
\begin{table}[H]
		\begin{center}
		\caption{Discriminator conventional architecture in CIFAR experiment }
		\begin{tabular}{|c c c|}
		    \hline
		    Layer & output size &filter \\
		    \hline
			Conv2D & $64\times32\times32$ & $3 \rightarrow 64$\\
            Conv2D & $64\times32\times32$ & $64 \rightarrow 64$\\
            Conv2D & $128\times32\times32$ & $64 \rightarrow 128$\\
			Avg-Pool2D & $128\times16\times16$ & -\\
            \hline
			Conv2D & $128\times16\times16$ & $128 \rightarrow 128$\\
            Conv2D & $256\times16\times16$ & $128 \rightarrow 256$\\
			Avg-Pool2D & $256\times8\times8$ & -\\
            \hline
			Conv2D & $256\times8\times8$ & $256 \rightarrow 256$\\
            Conv2D & $512\times8\times8$ & $256 \rightarrow 512$\\
			Avg-Pool2D & $512\times4\times4$ & -\\
            \hline
			Conv2D & $512\times4\times4$ & $512 \rightarrow 512$\\
            Conv2D & $512\times4\times4$ & $512 \rightarrow 512$\\
            \hline
            Reshape & $512\cdot4\cdot4$ & -\\
			Fully Connected & $1$ & $512\cdot4\cdot4\rightarrow1$\\
			\hline
		\end{tabular}
		\end{center}
	\end{table}

\begin{table}[H]
		\begin{center}
		\caption{Generator architecture in ImageNet experiment }
		\begin{tabular}{|c c c|}
		    \hline
		    Layer & output size &filter \\
		    \hline
			Fully connected & $1024 \cdot 4 \cdot 4$ & $256 \rightarrow 1024 \cdot 4\cdot4$\\
			Reshape & $1024\times4\times4$ & -\\
			\hline
			Resnet-Block & $1024\times4\times4$ & $1024\rightarrow1024\rightarrow1024$\\
            Resnet-Block & $1024\times4\times4$ & $1024\rightarrow1024\rightarrow1024$\\
			NN-Upsampling & $1024\times8\times8$ & -\\
			\hline
			Resnet-Block & $512\times8\times8$ & $1024\rightarrow512\rightarrow512$\\
            Resnet-Block & $512\times8\times8$ & $512\rightarrow512\rightarrow512$\\
			NN-Upsampling & $512\times16\times16$ & -\\
			\hline
			Resnet-Block & $256\times16\times16$ & $512\rightarrow256\rightarrow256$\\
            Resnet-Block & $256\times16\times16$ & $256\rightarrow256\rightarrow256$\\
			NN-Upsampling & $256\times32\times32$ & -\\
			\hline
			Resnet-Block & $128\times32\times32$ & $256\rightarrow128\rightarrow128$\\
            Resnet-Block & $128\times32\times32$ & $128\rightarrow128\rightarrow128$\\
			NN-Upsampling & $128\times64\times64$ & -\\
			\hline
			Resnet-Block & $64\times64\times64$ & $128\rightarrow64\rightarrow64$\\
            Resnet-Block & $64\times64\times64$ & $64\rightarrow64\rightarrow64$\\
			Conv2D & $3\times64\times64$ & $64\rightarrow3$\\
			\hline
		\end{tabular}
		\end{center}
	\end{table}
	
\begin{table}[H]
		\begin{center}
		\caption{Discriminator architecture in ImageNet experiment }
		\begin{tabular}{|c c c|}
		    \hline
		    Layer & output size &filter \\
		    \hline
			Conv2D & $64\times64\times64$ & $3 \rightarrow 64$\\
			\hline
			Resnet-Block & $64\times64\times64$ & $64\rightarrow64\rightarrow64$\\
            Resnet-Block & $128\times64\times64$ & $64\rightarrow64\rightarrow128$\\
			Avg-Pool2D & $128\times32\times32$ & -\\
\hline
			Resnet-Block & $128\times32\times32$ & $128\rightarrow128\rightarrow128$\\
            Resnet-Block & $256\times32\times32$ & $128\rightarrow128\rightarrow256$\\
			Avg-Pool2D & $256\times16\times16$ & -\\
			\hline
			Resnet-Block & $256\times16\times16$ & $256\rightarrow256\rightarrow256$\\
            Resnet-Block & $512\times16\times16$ & $256\rightarrow256\rightarrow512$\\
			Avg-Pool2D & $512\times8\times8$ & -\\
\hline
			Resnet-Block & $512\times8\times8$ & $512\rightarrow512\rightarrow512$\\
            Resnet-Block & $1024\times8\times8$ & $512\rightarrow512\rightarrow1024$\\
			Avg-Pool2D & $1024\times4\times4$ & -\\
\hline
			Resnet-Block & $1024\times4\times4$ & $1024\rightarrow1024\rightarrow1024$\\
            Resnet-Block & $1024\times4\times4$ & $1024\rightarrow1024\rightarrow1024$\\
			Fully Connected & $1$ & $1024\cdot4\cdot4\rightarrow1$\\
			\hline
		\end{tabular}
		\end{center}
	\end{table}

\section{Further results}
\label{app5}

\begin{figure}[H]
  \centering
  \subfigure[]{\includegraphics[width=1.3in]{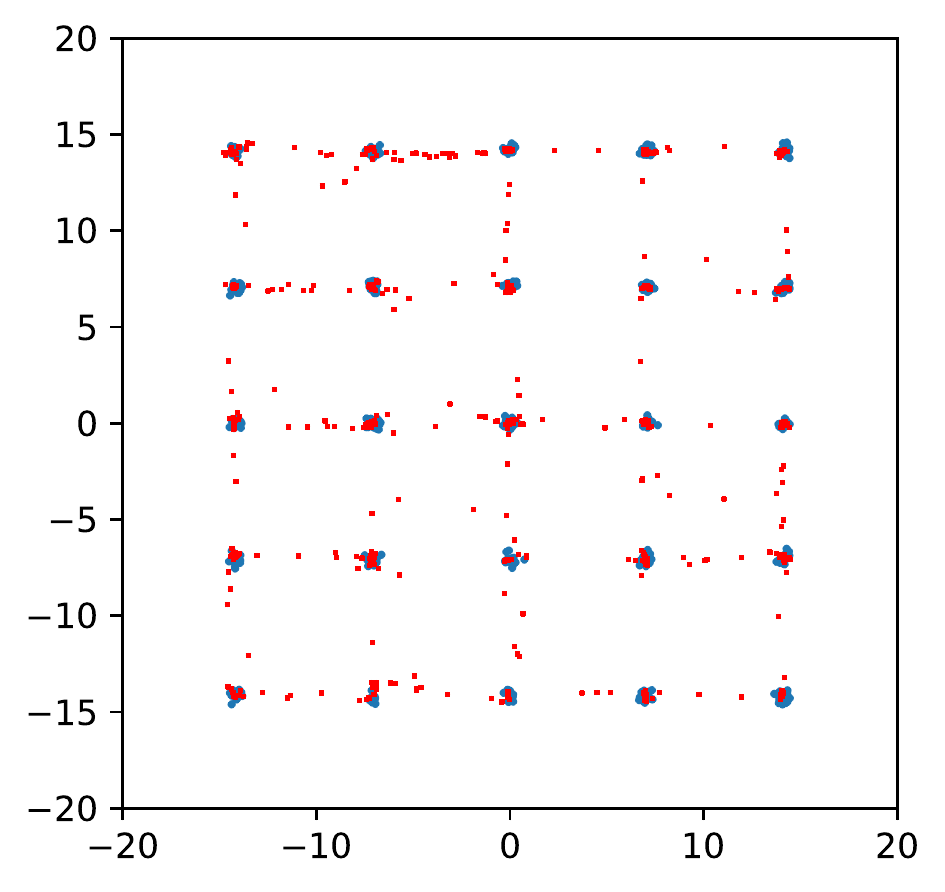}}
    \centering
  \subfigure[]{\includegraphics[width=1.3in]{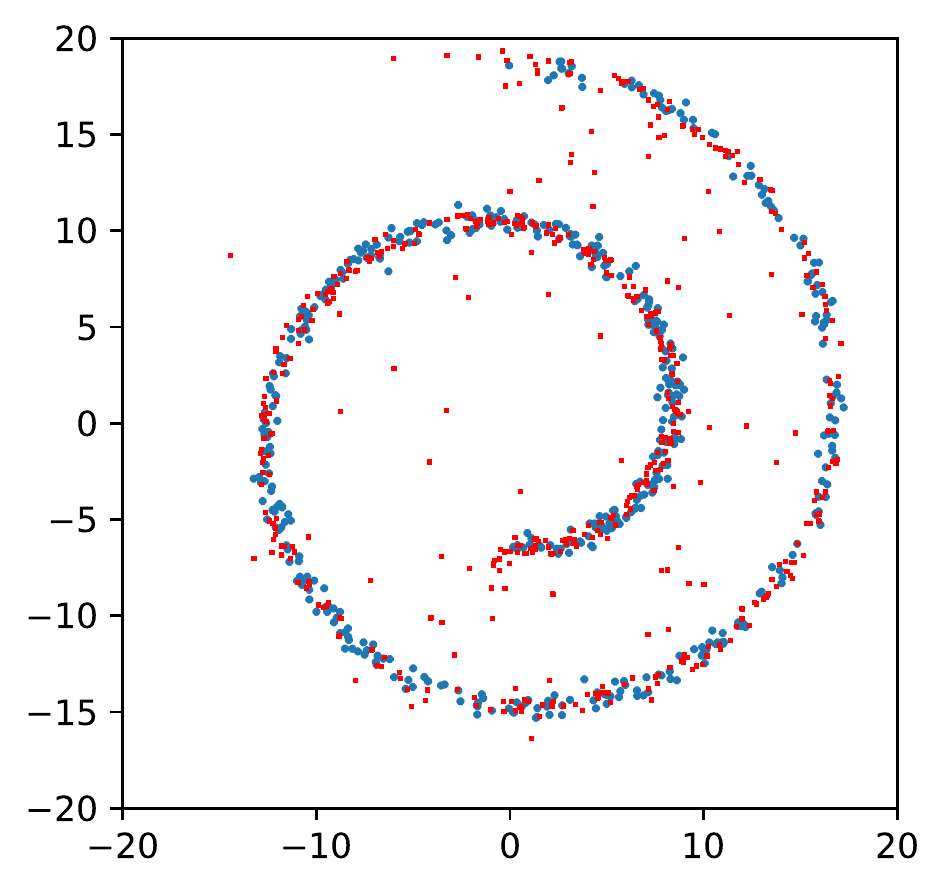}}
  \caption{Generation of our method on a mixture of $25$ Gaussians dataset and swissroll datatset. }
\end{figure}

\begin{figure}[H]
  \centering
  \includegraphics[width=2.6in,height=1.6in]{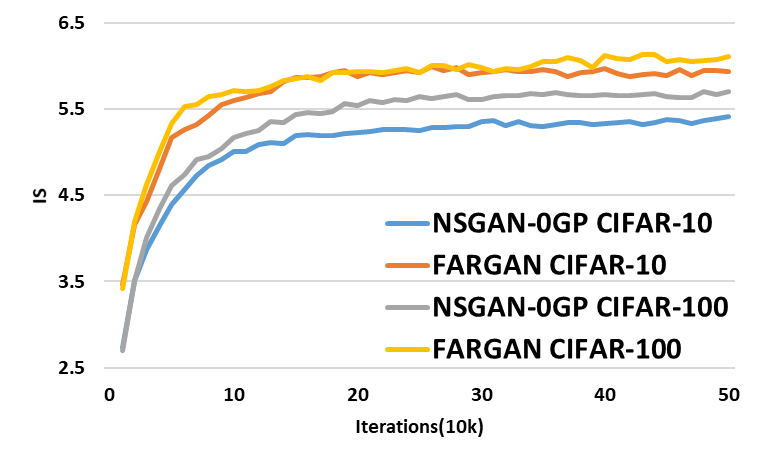}
  \caption{Inception score  on CIFAR-10 and CIFAR-100 of NSGAN-0GP and FARGAN for a DCGAN based network architecture. Our method still outperforms NSGAN-0GP.}
\end{figure}

\begin{figure}[H]
  \centering
  \subfigure[]{\includegraphics[width=2.6in,height=1.6in]{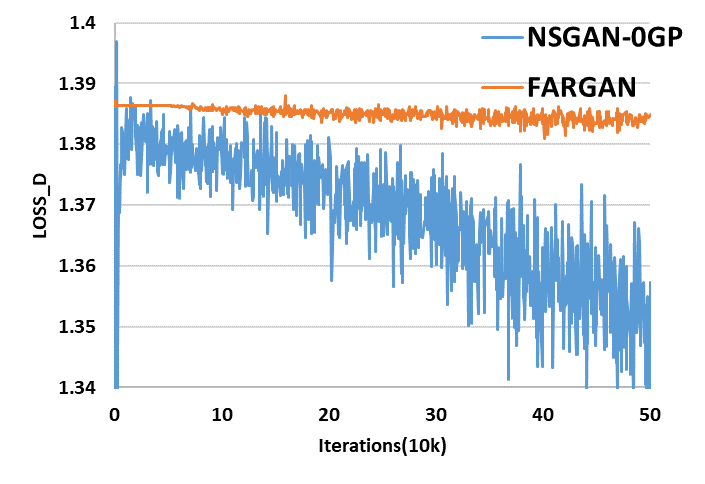}}
  \hspace{0.5cm}
    \centering
  \subfigure[]{\includegraphics[width=2.6in,height=1.6in]{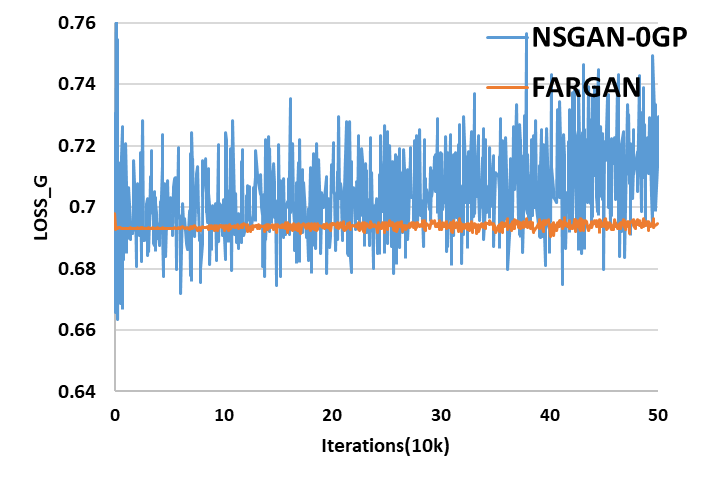}}
  \caption{Losses of discriminator (not including regularization term) and generator on CIFAR-100 of NSGAN-0GP and FARGAN}
\end{figure}

\begin{figure}[H]
  \centering
  \subfigure[]{\includegraphics[width=2.6in,height=1.6in]{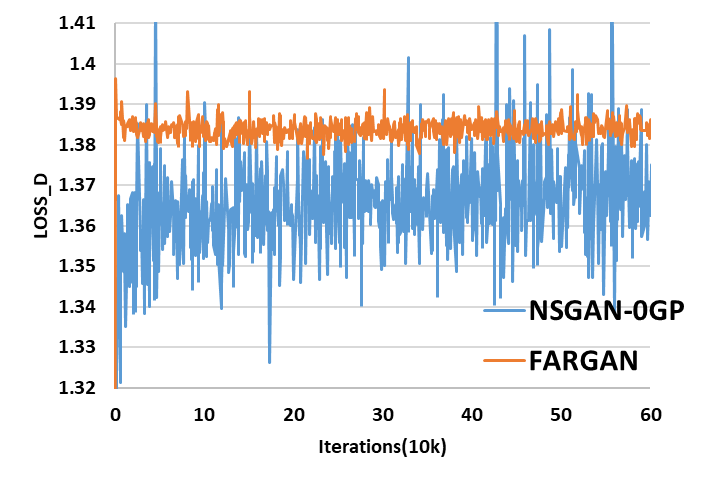}}
  \hspace{0.5cm}
    \centering
  \subfigure[]{\includegraphics[width=2.6in,height=1.6in]{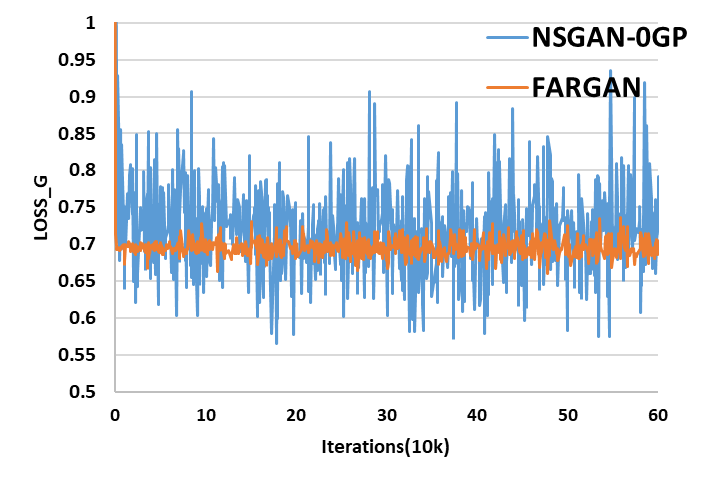}}
  \caption{Losses of discriminator (not including regularization term) and generator on ImageNet of NSGAN-0GP and FARGAN}
\end{figure}

\begin{figure}[H]
  \centering
  \subfigure[image generation of NSGAN-0GP]{\includegraphics[width=2.6in]{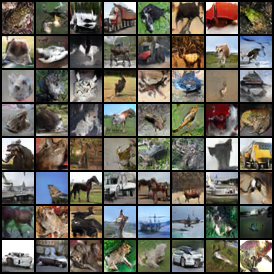}}
    \centering
  \subfigure[image generation of FARGAN]{\includegraphics[width=2.6in]{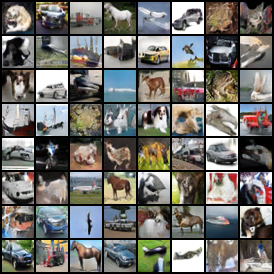}}
  \caption{Image generation of CIFAR-10.}
\end{figure}

\begin{figure}[H]
  \centering
  \subfigure[image generation of NSGAN-0GP]{\includegraphics[width=2.6in]{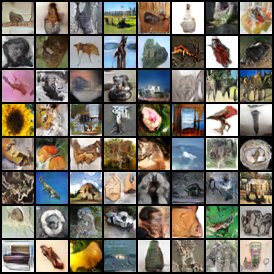}}
    \centering
  \subfigure[image generation of FARGAN]{\includegraphics[width=2.6in]{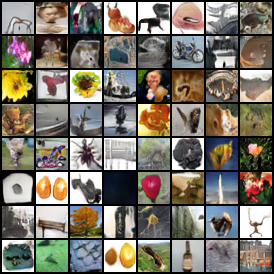}}
  \caption{Image generation of CIFAR-100.}
\end{figure}

\begin{figure*}
  \centering
  \subfigure[image generation of NSGAN-0GP]{\includegraphics[width=4in]{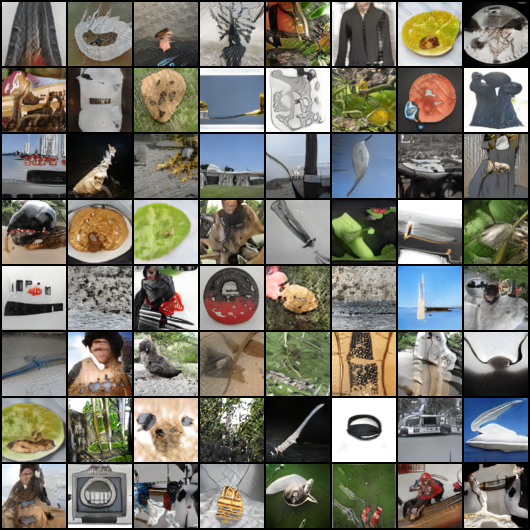}}
    \centering
  \subfigure[image generation of FARGAN]{\includegraphics[width=4in]{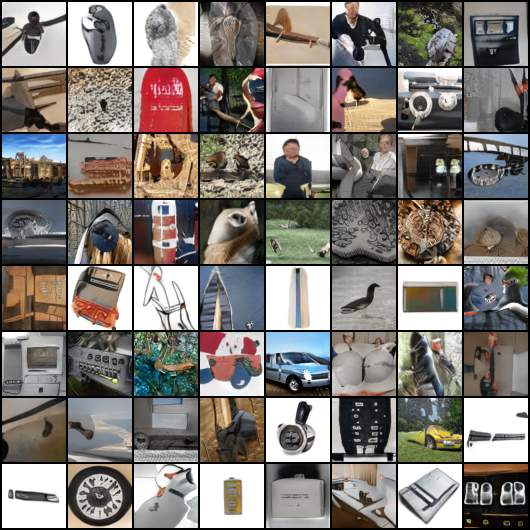}}
  \caption{Image generation of ImageNet.}
\end{figure*}

\end{document}